\documentclass[11pt,draftcls,onecolumn]{IEEEtran}
\usepackage{cite}
\usepackage{amsmath}
\usepackage{amssymb}
\usepackage{graphicx}
\usepackage{epstopdf}
\usepackage{subfigure}
\usepackage{epsfig}
\usepackage{algorithm}
\usepackage{algorithmic}

\begin{document}
%
% paper title
% can use linebreaks \\ within to get better formatting as desired
\title{Forest Sparsity for Multi-channel Compressive Sensing}
\author{Chen~Chen,~\IEEEmembership{Student Member,~IEEE,}
        Yeqing~Li,
        and~Junzhou~Huang*,~\IEEEmembership{Member,~IEEE}% <-this % stops a space
\thanks{Copyright (c) 2013 IEEE. Personal use of this material is permitted. However, permission to use this material for any other purposes must be obtained from the IEEE by sending a request to pubs-permissions@ieee.org.

Chen Chen, Yeqing Li and Junzhou Huang are with the Department of Computer Science and Engineering, University of Texas at Arlington, Texas, USA.
\emph{Asterisk indicates corresponding author}. Email: jzhuang@uta.edu.}
%of Electrical and Computer Engineering, Georgia Institute of Technology, Atlanta,
%GA, 30332 USA e-mail: (see http://www.michaelshell.org/contact.html).}% <-this % stops a space
%\thanks{J. Doe and J. Doe are with Anonymous University.}% <-this % stops a space
%\thanks{Manuscript received April 19, 2005; revised January 11, 2007.}}
%\author{Chen~Chen,
%        Yeqing~Li,
%        and~Junzhou~Huang*\\
%        Department of Computer Science and Engineering\\
%University of Texas at Arlington\\
%\texttt{\{cchen, yeqing.li\}@mavs.uta.edu} \\
%\texttt{jzhuang@uta.edu}\\
%EDICS:DSP-RECO
}

% If you want to put a publisher's ID mark on the page you can do it like
% this:
%\IEEEpubid{0000--0000/00\$00.00~\copyright~2007 IEEE}
% Remember, if you use this you must call \IEEEpubidadjcol in the second
% column for its text to clear the IEEEpubid mark.

% make the title area
\maketitle

\begin{abstract}

%Compressive sensing (CS) is an emerging techniquse to acquire sparse or compressible
%data with sub-Shannon/Nyquist sampling ratio.
In this paper, we investigate a new compressive sensing model for multi-channel sparse data where each channel can be represented as a hierarchical tree and different channels are highly correlated. Therefore, the full data could follow the forest structure and we call this property as \emph{forest
sparsity}. It exploits both intra- and inter- channel correlations and enriches the family of existing model-based compressive sensing theories. The proposed theory indicates that only $\mathcal{O}(Tk+\log(N/k))$ measurements are required for multi-channel data with forest sparsity, where $T$ is the number of channels, $N$ and $k$ are the length and sparsity number of each channel respectively. This result is much better than $\mathcal{O}(Tk+T\log(N/k))$ of tree sparsity,  $\mathcal{O}(Tk+k\log(N/k))$ of joint sparsity, and far better than  $\mathcal{O}(Tk+Tk\log(N/k))$ of standard sparsity. In addition, we extend the forest sparsity theory to the multiple measurement vectors problem, where the measurement matrix is a block-diagonal matrix. The result shows that the required measurement bound can be the same as that for dense random measurement matrix, when the data shares equal energy in each channel.
A new algorithm is developed and applied on four example applications to validate the benefit of the proposed model. Extensive experiments demonstrate the effectiveness and efficiency of the proposed theory and algorithm.

%which is much better than $\mathcal{O}(Tk+T\log(N/k))$ for tree sparsity , $\mathcal{O}(Tk+k\log(N/k))$, and far better than that of standard sparsity $\mathcal{O}(Tk+Tk\log(N/k))$.

% much less measurements are required for successful recovery in CS.

%In this paper, we investigate a new model called \emph{forest
%sparsity} for sparse learning and compressive sensing. It is an
%extension of standard sparsity when the support set of the data is consisted of a series of mutually correlated trees. Forest sparsity
%exists in many practical applications such as multi-contrast MRI,
%parallel MRI, multispectral image and color image recovery. We
%theoretically prove the benefit of forest sparsity, that much less
%measurements are required for successful recovery in compressive
%sensing. Moreover, a new algorithm is proposed and applied on
%several applications with forest sparsity. All experimental results
%validate the superiority of forest sparsity.

\end{abstract}

% Note that keywords are not normally used for peerreview papers.

\IEEEpeerreviewmaketitle

\section{Introduction}

\IEEEPARstart{S}{parsity} techniques are becoming more and more popular in
machine learning, statistics, medical imaging and computer vision as the emerging of compressive sensing. Based on compressive sensing theory \cite{candes2006robust,donoho2006compressed}, a small number of measurements are enough to recover the original data, which is an alternative to
Shannon/Nyquist sampling theorem for sparse or compressible data acquisition.

\subsection{Standard Sparsity and Algorithms}

Suppose $A  \in \mathbb{R}^{ M \times N}$ is the sampling matrix and $b
\in \mathbb{R}^M$ is the measurement vector, the problem is to recover the sparse data $x \in \mathbb{R}^N$ by solving the linear system $Ax = b$. Sometimes the data is not sparse but compressible under some base $\Phi$ such as wavelet, and the corresponding problem is $A \Phi^{-1} \theta = b$ where $\theta$ denotes the set of wavelet coefficients. Although the problem is underdetermined, the data can be perfectly reconstructed if the sampling matrix satisfy restricted isometry property (RIP) \cite{candes2006compressive} and the number of measurements is larger than $\mathcal{O}(k + k\log (N/k))$ for $k$-sparse data\footnote{We mean there are at most $k$ non-zero components in the data.} \cite{candes2007sparsity,candes2006stable}.

To solve the underdetermined problem, we may find the sparsest solution via $\ell_0$ norm regularization.  However, because the problem is NP-hard \cite{natarajan1995sparse} and impractical for most applications , $\ell_1$ norm regularization methods such as the lasso \cite{tibshirani1996regression} and basis pursuit (BP)
\cite{chen1998atomic} are first used to pursue the sparse solution. It has been proved that the $\ell_1$ norm regularization can exactly recover the sparse data for CS inverse problem under mild conditions \cite{donoho2003optimally,candes2006robust}. Therefore, a lot of efficient algorithms have been proposed for standard sparse recovery. Generally
speaking, those algorithms can be classified into three groups: greedy algorithms \cite{tropp2004greed,needell2009cosamp}, convex programming \cite{beck2009afast,figueiredo2007gradient,koh2007interior} and probability based methods \cite{ji2008bayesian,donoho2009message}.

%such as the orthogonal matching pursuit (OMP) \cite{tropp2004greed}, CoSaMP \cite{needell2009cosamp} and subspace pursuit (SP) \cite{dai2009subspace}; convex optimization algorithms that %contains thresholding \cite{beck2009afast}, reweighted least square \cite{gorodnitsky1997sparse}, gradient projection \cite{figueiredo2007gradient} and interior point method %\cite{koh2007interior}; Bayesian method \cite{ji2008bayesian} and approximate message passing (AMP) \cite{donoho2009message}.

\subsection{Joint Sparsity and Algorithms}

Beyond standard sparsity, the non-zeros components of $x$ often tend to be in some structures. This comes to the concept of \emph{structured sparsity} or model-based compressive sensing \cite{huang2011learning,baraniuk2010model,huang2011structured}. In contrast to standard sparsity that only relies on the sparseness of the data, structured sparsity models exploit both the non-zero values and the corresponding locations. For example, in the multiple measurement vector (MMV) problem, the data is consisted of several vectors that share the same support \footnote{The set of indices corresponding to the non-zero entries is often called the support}. This is called \emph{joint sparsity} that widely arise in cognitive radio networks \cite{meng2011collaborative}, direction-of-arrival estimation in radar \cite{krim1996two}, multi-channel compressive sensing \cite{baron2005distributed,majumdar2010compressive} and medical imaging \cite{bilgic2011multi,huang2012fast}. If the data $X\in \mathbb{R}^{TN \times 1}$ is consist of $T$ $k$-sparse vectors, the measurement bound could be substantially reduced to $\mathcal{O}(Tk + k\log (N/q))$ instead of $\mathcal{O}(Tk + Tk\log (N/q))$ for standard sparsity \cite{huang2010benefit,huang2011learning,baraniuk2010model,huang2009learningd}.

%if the data is consist of $T$ correlated vectors.s

A common way to implement joint sparsity in convex programming is to replace the  $\ell_1$ norm with $\ell_{2,1}$ norm, which is the summation of $\ell_2$ norms of the correlated entries \cite{yuan2005model,bach2008consistency}. $\ell_{2,1}$ norm for joint sparsity has been used in many convex solvers and algorithms \cite{cotter2005sparse,huang2012fast,van2008probing,deng2011group}. In Bayesian sparse learning or approximate message passing \cite{wipf2007empirical,ji2009multitask,ziniel2011efficient}, data from all channels contribute to the estimation of parameters or hidden variables in the sparse prior model.

%In AMP-MMV algorithm \cite{ziniel2011efficient}, the factor nodes for different measurement vectors share the same hidden variable, to indicate support set membership. %By geometric interpretation, $\ell_1$ norm
%ball has some singular values at the axes, which encourages sparseness. The singular values appears on $\ell_{2,1}$ norm ball only when all coordinates in the same group are zeros. %Intuitively, overlapping group inducing norm ball encourages overlapping group sparsity \cite{bach2011optimization}\cite{jacob2009group}.

%A lot of algorithms have gained significant benefit by exploiting these structured prior information. As an extension to lasso, the group lasso \cite{yuan2005model}\cite{koltchinskii2008sparse}\cite{obozinski2008high}\cite{bach2008consistency} replace the $\ell_1$ norm penalty with $\ell_{2,1}$ norm penalty, which is the summation of $\ell_2$ norms in group-wise. Similar method of utilizing $\ell_{2,1}$ norm for group sparsity has widely studied in convex programming algorithms and solvers \cite{cotter2005sparse}\cite{nie2010efficient}\cite{huang2012fast}\cite{van2008probing}\cite{wright2009sparse}\cite{liu2009slep}\cite{deng2011group}. In Bayesian sparse learning or Bayesian compressive sensing \cite{wipf2007empirical}\cite{ji2009multitask}, data from all sources contribute to the estimation of hyper-parameters in the sparse prior model.
\subsection{Tree Sparsity and Algorithms}

Another common structure would be the hierarchical tree structure, which has already been successfully utilized in image compression \cite{manduca1996wavelet}, compressed imaging \cite{he2009exploiting,som2012compressive,rao2011convex,chen2012compressive}, and machine learning \cite{kim2012tree}. Most nature signals/images are approximately tree-sparse under the wavelet basis. A typical relationship with \emph{tree sparsity} is that, if a node on the tree is non-zero, all of its ancestors leading to the root should be non-zeros. For multi-channel data $X= [x_1;x_2,...;x_T]\footnote{In this article, [;] denotes concatenating the data vetically.} \in \mathbb{R}^{NT\times 1}$, $\mathcal{O}(Tk + T\log (N/k))$ measurements are required if each channel $x_t$ is tree-sparse.

%Unlike group sparsity, the tree sparsity has been less studied due to its overlapping and intricate structure.

Due to the overlapping and intricate structure of tree sparsity, it is much harder to implement. For greedy algorithms, %Model-based CoSaMP \cite{baraniuk2010model}
%utilizes the condensing sort and select algorithm (CSSA) \cite{baraniuk2002near} to approximate the wavelet tree structure. Based on OMP \cite{tropp2007signal},
StructOMP \cite{huang2011learning} and TOMP \cite{la2006tree} are developed for exploiting tree structure where the coefficients are updated by only searching the subtree blocks instead of all subspace.  In statistical models \cite{he2009exploiting,som2012compressive}, hierarchical inference is used to model the tree structure, where the value of a node is not independent but relies on the distribution or state of its parent. In convex programming \cite{rao2011convex,chen2013benefit}, due to the tradeoff between the recovery accuracy and computational complexity, this is often approximated as overlapping group sparsity \cite{jacob2009group}, where each node and its parent are assigned into one group.

%Although $\ell_1$ norm regularization yields sparse solutions, the
%result could be even better if more priors can be utilized. The
%concept \emph{structured sparsity} \cite{baraniuk2010model}\cite{huang2011learning}\cite{bach2011structured}
%arises when data is not only sparse, but also organized in
%structures (e.g. groups, trees, graphs). A well known instance \emph{group
%sparsity} or \emph{block sparsity} \cite{baron2005distributed}\cite{yuan2005model} assumes the components of the data are zeros or
%nonzeros in group-wise. $\ell_{2,1}$ norm is often chosen as the
%group sparse penalty because it encourages the components in the
%same group to be zeros or non-zeros simultaneously. When one component
%appears in multiple groups, the problem becomes overlapping group
%sparse regularization \cite{jacob2009group}. Another example would be the tree sparsity,
%such as that the wavelet coefficients of a natural signal/image yield a
%binary tree/quadtree. The coefficients on the same subtree tend to be
%zeros or non-zeros simultaneously. Compared to the group sparsity,
%tree sparsity is often approximated by convex programming due to the
%difficulty of overlapping and hierarchical structure
%\cite{kim2009tree}\cite{zhao2009composite}\cite{jenatton2012multiscale}. Generally speaking, both the group
%sparsity and tree sparsity belong to \emph{graph sparsity}, where each
%components can be viewed as a vertex while the connections are
%edges. %There may be no fixed number of components in each group and
%the positions of the groups neither.

\subsection{Forest Sparsity}

Although both joint sparsity and tree sparsity have been widely studied, unfortunately, there is no work that study the benefit of their combinations so far. Actually, in many multi-channel compressive sensing or MMV problems, the data has joint sparsity across different channels and each channel itself is tree-sparse. Note that this differs from C-HiLasso \cite{sprechmann2011c}, where sparsity is assumed inside the groups. No method has fully exploited both priors and no theory guarantees the performance. In practical applications, researchers and engineers have to choose either joint sparsity algorithms by giving up their intra tree-sparse prior, or tree sparsity algorithms by ignoring their inter correlations.

In this paper, we propose a new sparsity model called \emph{forest
sparsity} to bridge this gap. It is a natural extension of existing structured sparsity models by
assuming that the data can be represented by a forest of mutually connected trees.
We give the mathematical definition of forest sparsity. Based on compressive sensing theory, we prove
that for a forest of $T$ k-sparse trees, only $\mathcal{O}(Tk+\log
(N/k))$ measurements are required for successful recovery with high
probability. That is much less than the bounds of joint sparsity
$\mathcal{O}(Tk + k\log (N/k))$ and tree sparsity $\mathcal{O}(Tk+T
\log (N/k))$ on the same data. The theory is further extended to the case on MMV problems, which is ignored in existing structured sparsity theories \cite{huang2011learning,baraniuk2010model,huang2011structured}. Finally, we derive an efficient algorithm to optimize
the forest sparsity model. The proposed algorithm is applied on
medical imaging applications such as multi-contrast magnetic resonance
imaging (MRI), parallel MRI (pMRI),  as well as color images, multispectral image reconstruction. % All experiments on real
%data of these issues show significant improvement to the
%state-of-the-art methods with forest sparse learning.
Extensive experiments demonstrate the advantages of forest sparsity
over the state-of-the-art methods in these applications.

\subsection{Paper Organization}

The remainder of the paper is organized as follows. Existing works for standard sparsity, joint sparsity and tree sparsity are reviewed in Section \ref{sec:rel}. We will propose forest sparsity and give the benefit of forest sparsity in Section \ref{sec:for}. An algorithm is presented in Section \ref{sec:alg}. We conduct experiments on four applications compared with standard sparsity, joint sparsity, and tree sparsity algorithms in Section \ref{sec:exp}. And finally the conclusion is drawn in Section \ref{sec:con}.

% The theoretical benefit of forest sparsity compared with standard sparsity, joint sparsity, and tree sparsity will be discussed in Section \ref{sec:the}.

\section{Background and Related Work}\label{sec:rel}
% A signal $x\in
%\mathbb{R}^{N}$ is $k$-sparse if only $k$ ($k \ll N$) entries of x are
%non-zeros. %The set of indices corresponding to the non-zero
%entries is often called the support of $x$ and denoted it by $supp(x)$. Thus,

In compressive sensing (CS), the capture of a sparse signal
 and compression are integrated into a single process
~\cite{donoho2006compressed,candes2006compressive}. We do not capture sparse data $x  \in \mathbb{R}^{N}$ directly but rather capture $M<N$ linear
measurements $b=A x$ based on a measurement matrix $A \in
\mathbb{R}^{M \times N}$. %Suppose the set of $k$-sparse signal $x$
%lives in the union $\Omega_{k}$ of $k$-dimensional subspaces, the
%union $\Omega_{k}$ thus includes $C_{N}^{k}$ subspaces.
To stably
recover the $k$-sparse data $x$ from $M$ measurements, the
measurement matrix $A$ is required to satisfy the Restricted
Isometry Property (RIP) ~\cite{candes2006compressive}. Let $\Omega_k$ denote the union $k$-dimensional subspaces where $x$ lives in.

\textbf{Definition 1: ($k$-RIP)} \emph{An $M \times N$  matrix $A$ has the $k$-restricted isometry property
with restricted isometry constant $1>\delta_k>0$, if for all $x  \in  \Omega_k$, and}
\begin{eqnarray}
(1-\delta_k)||x||_2^2 \leq ||Ax||_2^2 \leq (1+\delta_k)||x||_2^2.
\end{eqnarray}

%\emph{For any matrix $A\in
%\mathbb{R}^{M\times N}$ and any integer $k<M$, for all $x$ in the
%union $\Omega_k$, if there exists a constant $1>\delta_k>0$ and
%
%the matrix $A$ is said to satisfy the $k$-restricted isometry property
%with restricted isometry constant $\delta_k$}.

CS result shows that, if $M=\mathcal{O}(k+k\log(N/k))$, a sub-Gaussian random
matrix\footnote{It includes Gaussian and Bernoulli random matrices etc. \cite{mendelson2008uniform}.} $A$ can
satisfy the RIP with high probability \cite{baraniuk2008simple,blumensath2009sampling}.

Recently, structured sparsity theories demonstrate that when there is some structured prior information (e.g. group, tree, graph) in $x$, the measurement bound could be reduced \cite{baraniuk2010model,huang2011learning}. Suppose $x$ is in the union of
subspaces $\mathcal{A}$, then the $k$-RIP can be extended to the
$\mathcal{A}$-RIP \cite{blumensath2009sampling}:

%$\Omega_k$ includes all the $C_N^k$ subspaces if there is no further
%constrain on the support set of sparse data $x$. However, when $x$ has some
%structured sparsity property(e.g. tree sparsity) and

\textbf{Definition 2: ($\mathcal{A}$-RIP)} \emph{An $M \times N$  matrix $A$ has the $\mathcal{A}$-restricted isometry property
with restricted isometry constant $1>\delta_{\mathcal{A}}>0$, if for all $x  \in \mathcal{A}$, and}
\begin{eqnarray}
(1-\delta_{\mathcal{A}})||x||_2^2 \leq ||Ax||_2^2 \leq (1+\delta_{\mathcal{A}})||x||_2^2.
\end{eqnarray} \vspace{-0.0cm}
$\mathcal{A}$-RIP property has been proved to be sufficient for robust recovery of structured-sparse signals under noisy conditions \cite{baraniuk2010model}.
The required number of measurements $M$ has been quantified for a sub-Gaussian random matrix $A$ that has
the $\mathcal{A}$-RIP \cite{blumensath2009sampling}:

\textbf{Theorem 1: ($\mathcal{A}$-RIP)} \emph{Let $\mathcal{A}$ be
the union of $L$ subspaces of $k$ dimension in $\mathbb{R}^N$. For
any $t>0$, let
\begin{eqnarray}
M \geq \frac{2}{c\delta_{\mathcal{A}_k}}(\ln (2L) + k\ln \frac{12}{\delta_{\mathcal{A}_k}}+t), \label{eqn:mbound}
\end{eqnarray}
then there exists a constant $c>0$ and a randomly generated sub-Gaussian matrix  $A\in \mathbb{R}^{M\times N}$ satisfies
the $\mathcal{A}$-RIP with probability
at least $1-e^{-t}$}.

From (\ref{eqn:mbound}), we could intuitively observe that $M$ can
be less by reducing the number of subspaces $\mathcal{A}$. It
coincides with the intuition that the result will be improved when more priors are
utilized. For standard $k$-sparse data, there is no more constraint
to reduce the number of possible subspaces $C_N^k$. Let $L = C_N^k \approx (eN/k)^k$, the CS result for standard sparsity can be derived from Theorem 1.

%Therefore,

%\textbf{Corollary 1}: \emph{For $k$-sparse data, the sampling matrix
%$A$ has the $k$-RIP with probability $1-e^{-t}$ if the bound for the number of
%measurements satisfies that $M=\mathcal{O}(k+k\log(N/k))$.}
%\begin{eqnarray}
%m=\mathcal{O}(k+k\log(N/k))
% \label{eqn:Kbound}
%\end{eqnarray} \vspace{-0.0cm}}
%Actually, some data is not only sparse, but also structured formed. One type of such data is tree-sparse.
Now we consider structured sparse data.
Following \cite{baraniuk2010model}, if a $k$-sparse data
$x\in\mathbb{R}^N$ can form a tree or can be sparsely represented as
a tree under one orthogonal sparse basis $\Phi$ (e.g. wavelet), and the $k$
non-zero components naturally form a subtree, then it is called tree-sparse data.

%Let $\cdot(j)$ denotes the $j$-th entry of a vector and $parent(\cdot)$ denotes the parent coefficient.

\textbf{Definition 3:} \emph{Tree-sparse data in $\mathbb{R}^N$ is defined as} \\ \emph{$\mathcal{T}_k$=\{$x=\Phi^{-1}\theta$:
$\theta|\Omega^C=0$, $|\Omega|=k$, where $\Omega$ forms a connected subtree.  \}.}
\vspace{0.1cm}
%}

Here $\Omega \subseteq \{1,2,...,N\}$ denotes a subspace of the data as and the support is in $\Omega$.
$\Omega^C$ denotes the complement of $\Omega$ and $\theta$ denotes the coefficients under $\Phi$. It implies that, if an entry of $\theta$ is in $\Omega$ , all its ancestors on the tree must be in $\Omega$.

For tree-sparse data, we say it has the \emph{tree sparsity} property.
Most natural signals or images have tree sparsity property, since they can be
sparsely represented with the wavelet tree structure. Specially, the wavelet
coefficients of a 1D signal form a binary tree and those of a 2D image yield a quadtree. If the union of all subspaces are denoted by $\Omega_{Tree}$, it is obviously that $\Omega_{Tree} \subset \Omega_{k}$ and the number of subspaces $L_{Tree}<C_N^k$. %L_k$.

\textbf{Theorem 2}: \emph{For tree-sparse data, there exists a sub-Gaussian random matrix $A\in
\mathbb{R}^{M \times N}$ that has the $\mathcal{T}_{k}$-RIP with probability $1-e^{-t}$ if the number of
measurements satisfies that:}
\begin{eqnarray}
M \geq \left\{\begin{matrix}
\ \frac{2}{c_1\delta_{\mathcal{T}_{k}}} ( k + \ln(N/(k+1)) + k\ln(12/\delta_{\mathcal{T}_{k}})  \\ + \ln 2+t) \qquad \textrm{if} \ k \leq \left \lfloor \log_2 N \right \rfloor , \\
\ \frac{2}{c_1\delta_{\mathcal{T}_{k}}} (k\ln4 + \ln(c_2 N/k) + k\ln(12/\delta_{\mathcal{T}_{k}})
 \\ + \ln 2+t) \qquad \textrm{if} \ k>\left \lfloor \log_2 N \right \rfloor .
\end{matrix}\right.
\end{eqnarray}
\emph{where $c_1$ and $c_2$ denote absolute constants}.

For both case, we have $M=\mathcal{O}(k+\log(N/k))$. Similar conclusion has been drawn in previous articles \cite{huang2011learning}\cite{baraniuk2010model}. %Both the above models are for single channel data.%\begin{eqnarray}
%\label{eqn:treebound}
%\end{eqnarray} \vspace{-0.0cm}}

So far, we have reviewed standard sparsity and tree sparsity on single channel data.
For multi-channel data that contains $T$ channels or vectors (i.e. $X=[x_1;x_2;...;x_T] \in \mathbb{R}^{NT\times 1}$), each of which is standardly $k$-sparse, the bound for the number of measurement should be $\mathcal{O}(Tk+Tk\log(N/k))$. If each channel is tree-sparse and independently, the measurement bound for a sub-Gaussian random matrix $A\in \mathbb{R}^{TM\times TN}$
is $TM=\mathcal{O}(Tk+T\log(N/k))$.

%\textbf{Lemma 2}: \emph{For T-channel data, each channel is tree-sparse and independent, a
%subgaussian random matrix $A$ has the $\mathcal{T}_{T,k}$-RIP with
%probability $1-e^{-t}$ if the number of measurements satisfies that:}
%\begin{eqnarray}
%M \geq \left\{\begin{matrix}
%\ \frac{2}{c\delta_{\mathcal{T}_{T,k}}} ( Tk + T\ln(N/(k+1)) + Tk\ln(12/\delta_{\mathcal{T}_{T,k}})  \\ + \ln 2+t) \qquad \textrm{if} \ k \leq \left \lfloor \log_2 N \right \rfloor , \\
%\ \frac{2}{c\delta_{\mathcal{T}_{T,k}}} ( Tk\ln4 + T\ln(c'N/k) + Tk\ln(12/\delta_{\mathcal{T}_{T,k}})
% \\ + \ln2  +t) \qquad \textrm{if} \ k>\left \lfloor \log_2 N \right \rfloor .
%\end{matrix}\right.  \notag
%\end{eqnarray} \vspace{-0.0cm}
%For both case, we have $M=\mathcal{O}(Tk+T\log(N/k))$.
It is important to note that the T-channel $k$-sparse data has sparsity $Tk$ but not $k$.
Different from the above independent channels, another case is that all channels of the data may be highly correlated, which corresponds to joint sparse data:%. If each channel share the same support, the data is called joint-sparse data:

\textbf{Definition 4:} \emph{Joint-sparse data is defined as }\\ \emph{$\mathcal{J}_{T,k}$=\{$X=[x_1;x_2;...;x_T]$: $x_i=\Phi^{-1}\theta_i$,
$\theta_i|\Omega^C=0$, $|\Omega|=k$, $i=1,2,...,T$ \} }.
\vspace{0.1cm}

Similar as tree-sparse data, joint-sparse data has the \emph{joint sparsity} property. It has to be clarified that joint sparsity does not rely on tree sparsity. The former utilizes the structure across different channels, while the later utilizes the structure within each channel. Previous works implies that the minimum measurement bound for such joint sparse data is $TM=\mathcal{O}(Tk+k\log(N/k))$ \cite{huang2010benefit,huang2011learning,baraniuk2010model}.

%\textbf{Lemma 3}: \emph{For joint-sparse data, a subgaussian random matrix $A$
%has the $\mathcal{J}_{T,k}$-RIP with probability $1-e^{-t}$ if the number of
%measurements satisfies that:}
%\begin{eqnarray}
%M \geq
%\ \frac{2}{c\delta_{\mathcal{J}_{T,k}}} [\ln 2 + k\ln(eN/k)
%+ Tk\ln(12/\delta_{\mathcal{J}_{T,k}}) +t]
%\notag
%\end{eqnarray} \vspace{-0.0cm}
%%\begin{eqarray}
%M=\mathcal{O}(Tk+k\log(N/k))
%\label{eqn:Groupbound}
%\end{eqnarray} \vspace{-0.0cm}}

%It implies that the minimum measurement bound is $M=\mathcal{O}(Tk+k\log(N/k))$.
%This result has been proved in previous works \cite{huang2010benefit}\cite{huang2011learning}\cite{baraniuk2010model}. Compared the results of Lemma 2 and Lemma 3, it is hard to say which one is better, which depends on $T,k$ and how well the data follow the joint-sparse or tree-sparse structure.

%If each channel follows the tree structure and recover them with independent tree sparsity, the measurements bound should be lower because more prior information is exploited.

%For forest sparse data that contains $T$ trees, the bound for the number of measurement should be $T\times \mathcal{O}(k+k\log(N/k))$ if they are recovered with standard sparsity, which is
%at least $\mathcal{O}(Tk+Tk\log(N/k))$. Instead, some methods model the data with joint sparsity or block sparsity \cite{baron2005distributed}\cite{deng2011group} by assuming the data in all $T$ channels share the same support set.

%Both joint sparsity and tree sparsity are not perfect on forest sparse data.

\section{Forest Sparsity} \label{sec:for}

\begin{figure*}[htbp]
\centering \vspace{-0.0cm}
    \subfigure[]{\label{fig:brain}
        \includegraphics[scale=0.21]{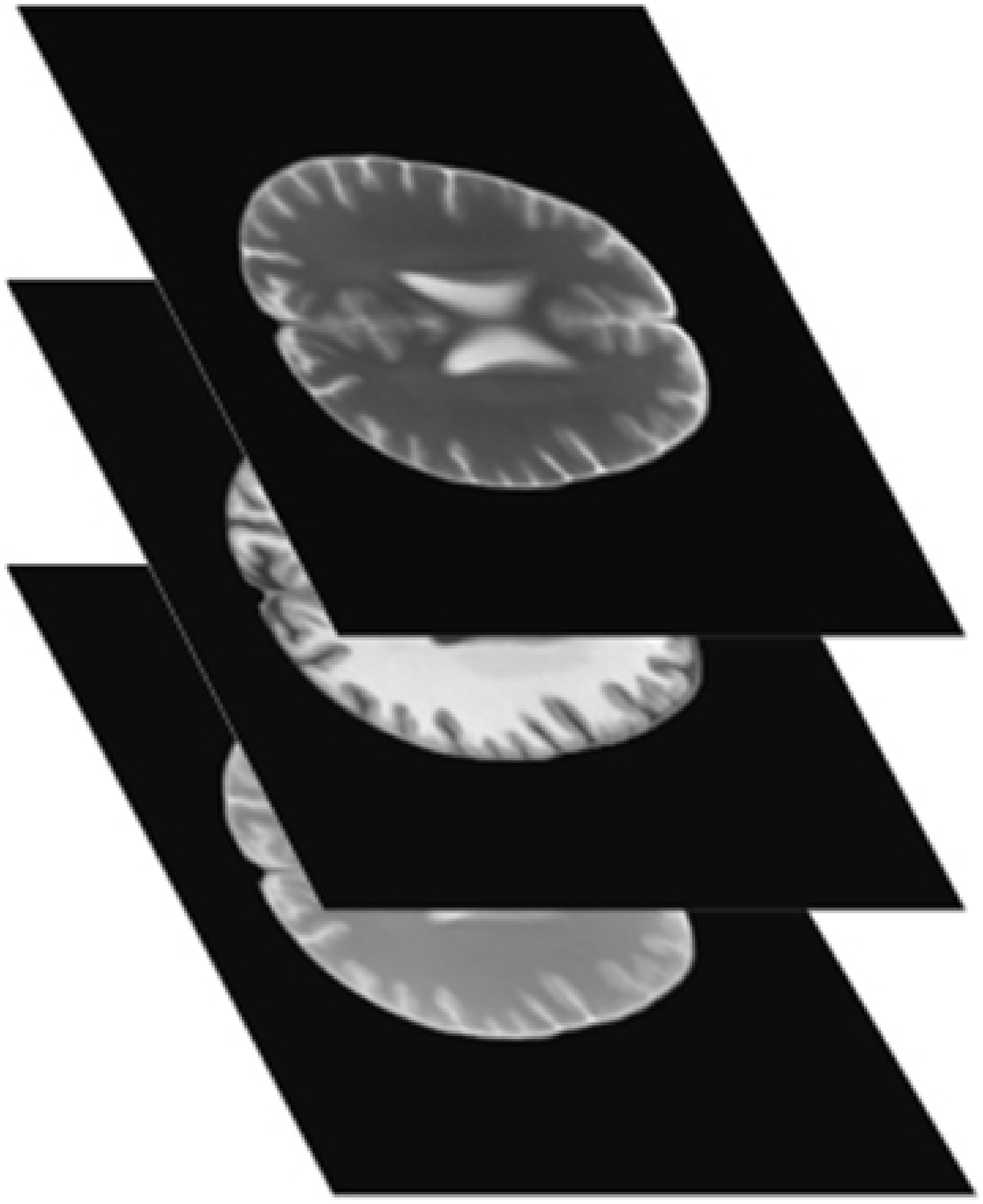}}
    \subfigure[]{\label{fig:tree}
        \includegraphics[scale=0.25]{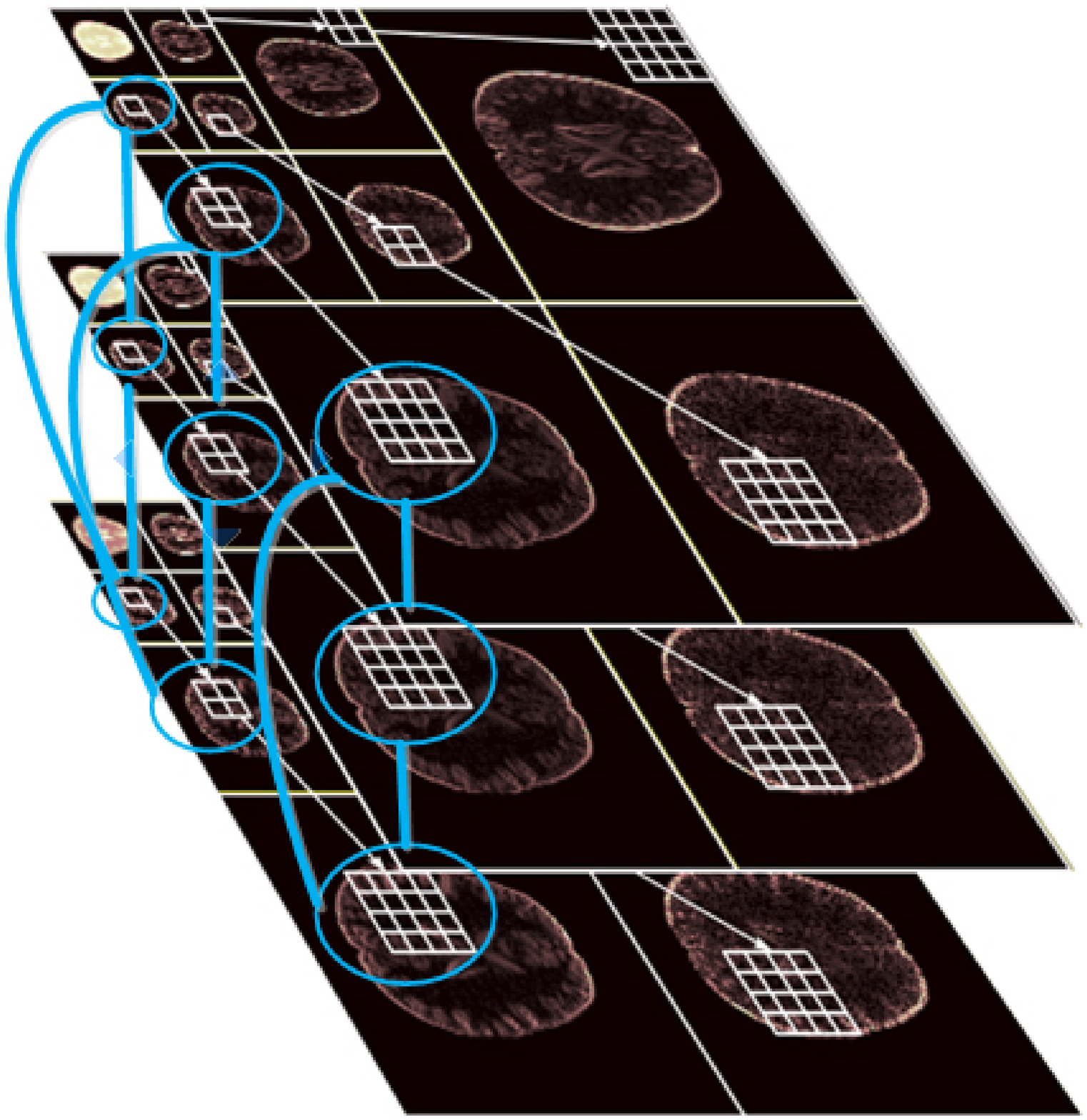}}
    \subfigure[]{\label{fig:brain}
        \includegraphics[scale=0.21]{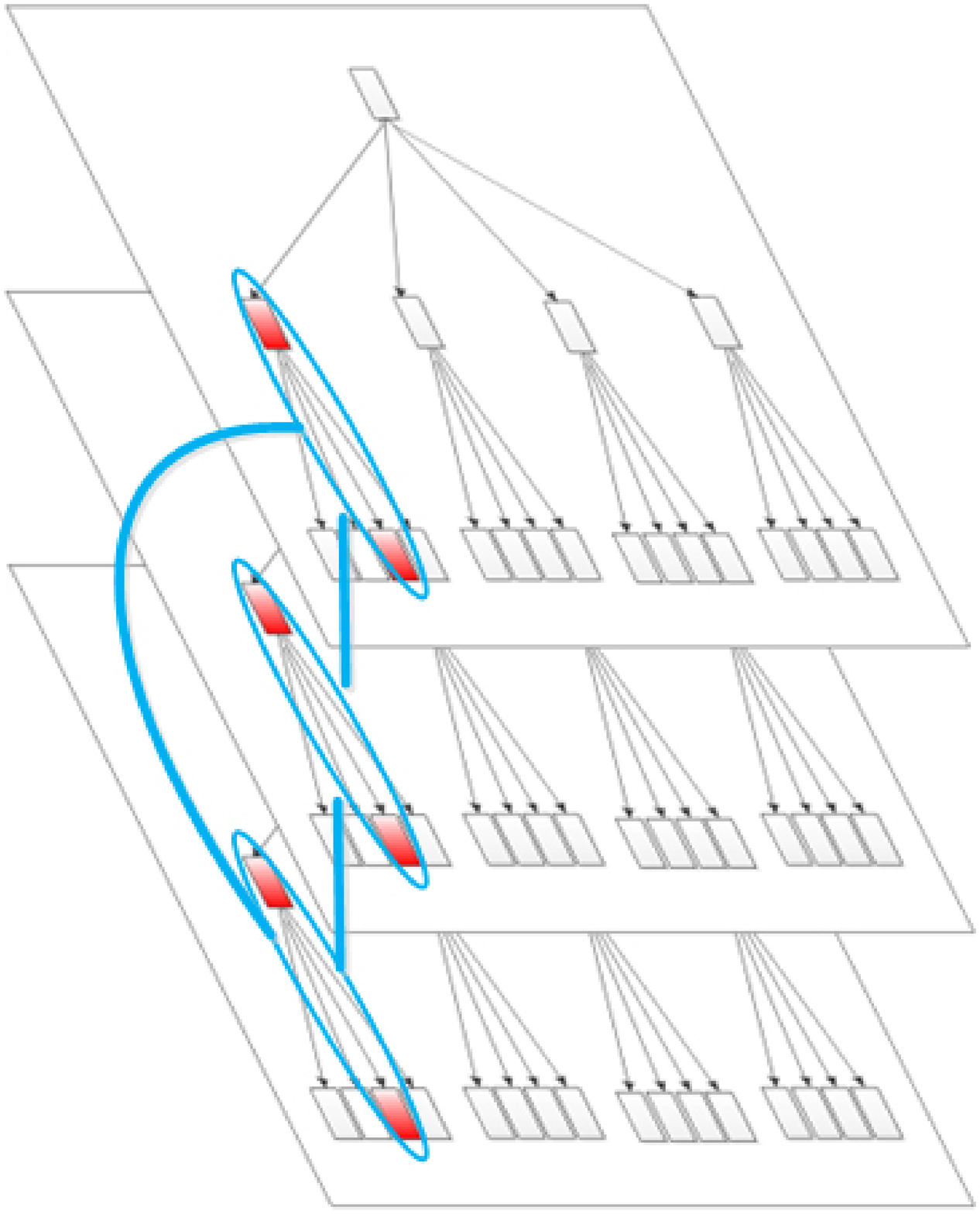}}
\caption{Forest structure on multi-contrast MR images. (a) Three
multi-contrast MR images. (b) The wavelet coefficients of the
images. Each coefficient tends to be consistent with its parent and
children, and the coefficients across different trees at the same
position. (c) One joint parent-child group across different trees that used
in our algorithm.}\label{fig:mtforest}\vspace{-0.0cm}
\end{figure*}

%As mentioned above, both tree sparsity and forest sparsity are
%extensions of standard sparsity.  They require that the data is not only
%sparse, but also follows some special structure.
In practical applications, it happens usually that multi-channel images, such as color images, multispectral images and MR images, have the joint sparsity and tree sparsity simultaneously.   It is because: (a) the wavelet coefficients of each channel naturally yield a quadtree; (b) all channels represent the same physical objects (e.g. nature scenes or human organs), and the wavelet coefficients of each channel tend to be large/small simultaneously due to same boundaries of the objects.
Therefore, the support of such data is consist of several connected trees and like a forest. Fig. \ref{fig:mtforest} shows the forest structure in multi-contrast MR images. We could find that the non-zero coefficients are not random distributed but forms a connected forest. Unfortunately, existing tree-based algorithms can only recover multi-channel data channel-by-channel separately, and it is unknown how to model the tree structure in existing joint sparsity algorithms. In addition, there are no theoretical results in previous works showing how much better the recovery can be improved by fully exploiting the prior information.

\vspace{-0.0cm}
 \vspace{-0.0cm}

Motivated by this limitation, we extend previous works to a more
special but widely existed case. For multi-channel data, if it is jointly sparse, and more importantly, the common support of different channels yields a subtree structure,
we call this kind of data forest-sparse data:

% \emph{forest sparsity}.  A forest in this context is
%considered as a set of fully connected trees but not individual
%ones. Each tree in the forest has the same number of nodes, edges
%and the same structure of the support set. More importantly, all the nodes at the same
%position across different trees are mutually connected, that is, the
%corresponding values of these nodes are zeros or nonzeros simultaneously.
%Here and later, a set of nodes are said to
%be consistent when their values are zeros or nonzeros
%simultaneously.
%

%If there are $T$ sparse trees in a forest, we can define the forest sparse data
%like the definition of tree sparse data.

\textbf{Definition 5:} Forest-sparse data is defined as \\ \emph{$\mathcal{F}_{T,k}$=\{$X=[x_1;x_2;...;x_T]$: $x_i=\Phi^{-1}\theta_i$,
$\theta_i|\Omega^C=0$, $|\Omega|=k$, where $\Omega$ forms a connected subtree,  $i=1,2,...,T$ \} }.
\vspace{0.1cm}
%\textbf{Definition: (Forest sparse data)} \emph{ If there are a
%series of trees that share a common size and structure, and the
%components at the same position across different trees are mutually
%connected, they are defined as a forest. A data is forest sparse only
%if it is sparse and its support set yields a subforest. That is,
%$\mathcal{F}_{T,k}$=\{$x_i=\Phi^{-1}\theta_i$, $i=1,2,...,T$,:
%$\theta|\Omega^C=0$,  $|\Omega|=Tk$\}, where $\Omega$ forms a
%subforest.}

Similarly, the forest-sparse data has \emph{forest sparsity} property. This definition implies that if the coefficients at the same position across different channels are non-zeros, all their ancestor coefficients are all non-zeros. Learning with forest sparsity, we search the sparsest solution that follow the forest structure in the CS inverse problem. Any solution that violates the assumption will be penalized. Intuitively, the solution will be more accurate. We obtain our main result in the following theorem:
%
%Different from them, our forest sparsity model can significantly reduce the measurement bound. By the definition of forest sparsity, when the positions
%of the support set for one tree are fixed, the positions of the supports
%on all other trees will be also fixed. Therefore, the freedom of
%subspaces is further limited and the number of measurements obeys:

\textbf{Theorem 3}:\emph{ For forest-sparse data, there exists a sub-Gaussian random matrix $A\in
\mathbb{R}^{TM \times TN}$ that has the $\mathcal{F}_{T,k}$-RIP with
probability $1-e^{-t}$ if the number of measurements satisfies that:}
\begin{eqnarray}
TM \geq \left\{\begin{matrix}
\ \frac{2}{c_1 \delta_{\mathcal{F}_{T,k}}} (  k + \ln(N/(k+1))
+ Tk\ln(12/\delta_{\mathcal{F}_{T,k}})  \\ + \ln 2 +t)  \qquad \textrm{if} \ k \leq \left \lfloor \log_2 N
\right \rfloor ,\\
\ \frac{2}{c_1 \delta_{\mathcal{F}_{T,k}}} ( k\ln4 + \ln((c_2 N)/k)
+ Tk\ln(12/\delta_{\mathcal{F}_{T,k}}) \\ + \ln2 +t)
  \qquad \textrm{if} \ k>\left \lfloor \log_2 N \right \rfloor .
\end{matrix}\right.
\end{eqnarray}
\emph{where $c_1$ and $c_2$ are absolute constants}.

For both cases, the bound is reduced to $ M =\mathcal{O}(Tk+\log(N/k))$.
%\begin{eqnarray}
%M=\mathcal{O}(Tk+\log(N/k))
%\label{eqn:Fbound}
%\end{eqnarray} \vspace{-0.0cm}}
The proofs of Lemma 1 as well as Lemma 2, 4 are included in the appendices. Using the $\mathcal{F}_{T,k}$-RIP, forest-sparse data can be robustly recovered from noisy
compressive measurements.

Table \ref{bounds} lists all the measurement bounds for the forest-sparse data with different models. Standard sparsity model only exploits the sparseness while no prior information about the locations of the non-zero elements is involved. It is the classical but worst model for forest-sparse data.  These location priors are partially utilized in joint sparsity and tree sparsity models. One of them only studies the correlations across channels, while the other one only learns the intra structure. Our result is significantly better than those of joint sparsity and tree sparsity, and far better than that of standard sparsity, especially when $N/k$ is large. Only the proposed model fully exploits all these structures.

%Modeling with
%standard sparsity requires the most measurements as no more
%structured prior are utilized. It's hard to say which one is better
%between group sparsity and tree sparsity, which depends on $T$
%and how well the data follow the jointly sparse or tree sparse
%structure. Forest sparsity outperforms all others if the data is
%forest sparse.
\begin{table}[h!]
\caption{Measurement bounds for forest-sparse data} \label{bounds}
\begin{center}
\newsavebox{\tbox}
\begin{lrbox}{\tbox}
\begin{tabular}{cl}
\hline
\multicolumn{1}{c}{Sparse Models} &\multicolumn{1}{c}{Measurement
Bounds}
\\
\hline
Standard Sparsity     &$\mathcal{O}(Tk+Tk \log
(N/k))$    \\
Joint Sparsity      &$\mathcal{O}(Tk+k \log
(N/k))$   \\
Tree Sparsity        &$\mathcal{O}(Tk+T \log
(N/k))$    \\
Forest Sparsity     &$\mathcal{O}(Tk+ \log
(N/k))$   \\
\hline \\
\end{tabular}
\end{lrbox}
\scalebox{1.2}{\usebox{\tbox}}
\end{center}
\end{table}
\vspace{-0.5cm}
%\section{Applications of Forest Sparsity}

So far, we have analyzed the result by forest sparsity over previous results. %It suggests that the measurement matrix $A$ requires less rows than that required in joint sparsity and tree sparsity, to preserving successful recovery with the same probability.
In all these results, the measurement matrix $A$ is assumed to be a dense sub-Gaussian matrix. However, in many practical problems, each data channel $x_t \in \mathbb{R}^{N}$ is measured by a distinct compressive matrix $A'_t \in \mathbb{R}^{M \times N}$, $t=1,2,...,T$, which are called multiple measurement vectors (MMV) problems or multi-task learning ( e.g., \cite{baron2005distributed,chen2010graph,chen2012two}). Here and later, we assume that $\{A'_t\}_{t=1}^T$ follow the same distribution but may be different. Therefore, the matrix $A$ is actually a block-diagonal matrix rather than a dense matrix. The linear system $b = Ax$ can be written as:
\begin{eqnarray}
\begin{bmatrix}
b_1\\
b_2\\
...\\
b_T
\end{bmatrix} = \begin{bmatrix}
A'_1 &  &  & \\
 & A'_2 &  & \\
 &  & ... & \\
 &  &  & A'_T
\end{bmatrix}\begin{bmatrix}
x_1\\
x_2\\
...\\
x_T
\end{bmatrix} \label{eqn:bigA}
\end{eqnarray}
The non-diagonal blocks in $A$ are all zeros. Intuitively, such block-diagonal matrices have no better results than the dense matrices that discussed above, due to the less randomness. Unfortunately, the performance of the random block-diagonal matrices has not been analyzed on structured sparse data before, as all existing structured sparsity theories concentrate on the dense random matrix \cite{huang2011learning,baraniuk2010model,huang2011structured}.
In this article, we extend the theoretical result to the block-diagonal matrix in the MMV problems.

\textbf{Theorem 4}:\emph{ For forest-sparse data, there exists a block-diagonal matrix $A$ composed by sub-Gaussian random matrices $\{A'_t\}_{t=1}^T$ as in (\ref{eqn:bigA}), that has the $\mathcal{F}_{T,k}$-RIP with
probability $1-e^{-t}$ if the number of measurements satisfies that:}
\begin{eqnarray}
TM \geq \left\{\begin{matrix}
\ \frac{2T}{c_1W}(\ln 2 + \ln(N/(k+1))  + Tk\ln(12/\delta_{\mathcal{F}_{T,k}}) \\  +  k +t),    \qquad \textrm{if} \ k \leq \left \lfloor \log_2 N
\right \rfloor ,\\
\ \frac{2T}{c_1W} (\ln2  + \ln((c_3 N)/k) + Tk\ln(12/\delta_{\mathcal{F}_{T,k}}) \\ + k\ln4 +t),
\qquad \textrm{if} \ k>\left \lfloor \log_2 N \right \rfloor .
\end{matrix}\right.
\end{eqnarray}
\emph{where $W=\min(c_2^2\delta_{\mathcal{F}_{T,k}}^2 \Gamma_2, c_2\delta_{\mathcal{F}_{T,k}}\Gamma_\infty)$}; $\Gamma_2 = \frac{(\sum_{t=1}^T ||x_t||^2_2 )^2 }{\sum_{t=1}^T ||x_t||^4_2}$ and $\Gamma_\infty = \frac{\sum_{t=1}^T ||x_t||^2_2 }{\max_{t=1}^T ||x_t||^2_2}$; \emph{$c_1$, $c_2$ and $c_3$ are absolute constants.} \vspace{0.1cm}

For both cases, the bound can be written as $ TM =\mathcal{O}(\frac{T^2k+T\log(N/k)}{\min(\Gamma_2, \Gamma_\infty)})$. In contrast to previous results on dense matrices with i.i.d sub-Gaussian entries, this bound also depends on the energy of the data. It is not hard to find that $1 \leq \Gamma_2 \leq T$ and $1 \leq \Gamma_\infty \leq T$. In the best case, when  $||x_1||_2 = ||x_2||_2 = ...= ||x_T||_2$ and $\Gamma_2 = \Gamma_\infty = T$, the measurement bound is $TM =\mathcal{O}({Tk+\log(N/k)}$. It shows a similar performance as the dense sub-Gaussian matrix in Theorem 3. In the worst case, the energy of the data concentrate on one single channel/task, i.e., all $||x_t||_2=0$ except a single index $||x_{t'}||_2\neq 0$. The measurement bound then is $TM =\mathcal{O}({T^2k+T\log(N/k)}$, which is even worse than that in Theorem 2 for independent tree sparse channels. Even for the same block-diagonal matrix, the analysis makes clear that its performance may varies significantly depending on the data being measured. In the worst case, their measurement bound can increase $T$ times. %Comparing Theorem 3 and 4, the results differ from the factor $T/\min(\Gamma_2, \Gamma_\infty)$.
However, the increased factor $T/\min(c_2^2\delta^2 \Gamma_2, c_2\delta \Gamma_\infty)$ for block-diagonal matrices also applies to standard sparse data, joint sparse data and tree sparse data. For the same measurement matrix and the same data, the advantage of forest sparsity still exists.
Due to this reason, we do not evaluate the term $\min(c_2^2\delta^2 \Gamma_2, c_2\delta \Gamma_\infty)$ in the experiments, while focus our interest on comparing different sparsity models on the same data.

% However, the worst case rarely happens in most practical problems. Similar, in the best case, the diagonal block sub-Gaussian matrices can achieve the same bounds as dense matrices for standard sparse data, joint sparse data and tree sparse data, respectively (shown in Table \ref{bounds}). In the worst case, their measurement bounds are increased by a factor of $T$.

%We summarize these discussions in the following corollary.

%\textbf{Corollary}

\section{Algorithm} \label{sec:alg}

In this article, the forest structure is approximated as overlapping group sparsity \cite{jacob2009group} with mixed $\ell_{2,1}$ norm. Although it may not be the best approximation, it is enough to demonstrate the benefit of forest sparsity. To evaluate the forest sparsity model, we need to compare different models via a similar framework.
From the definition of forest-sparse data, we could find that a coefficient is large/small, its parent and "neighbors"\footnote{Parent denotes the parent node on the same channel while neighbors mean coefficients at the same position on other channels.} also tend to be large/small. All parent-child pairs in the same position across different channels are assigned into one group, and the problem becomes overlapping group sparsity regularization. Similar scheme has been used in approximating tree sparsity \cite{rao2011convex,chen2012compressive}, where each node and its parent are assigned into one group.
%Let $||\Phi x||_{\mathcal{F},T} \approx \sum_{g \in \mathcal{G}} ||(\Phi x)_{g}||_2$, where $g$ denotes one of the coefficient groups discussed above (an example is demonstrated in Fig.1(c)), $(\cdot)_g$ denotes the coefficients in group $g$ and $\mathcal{G}$ is the set of all groups.
We write the approximated problem as:
\begin{eqnarray}
\min_x \frac{1}{2}||Ax-b||_2^2 + \lambda \sum_{g \in \mathcal{G}} ||(\Phi x)_g||_{2} \label{eqn:appro1}
\end{eqnarray}
where $g$ denotes one of the coefficient groups discussed above (an example is demonstrated in Fig.1(c)), $(\cdot)_g$ denotes the coefficients in group $g$ and $\mathcal{G}$ is the set of all groups.

The mixed $\ell_{2,1}$ norm encourages all the components in the
same group $g$ to be zeros or non-zeros simultaneously. With our
group configuration, it encourages forest sparsity. We present an
efficient implementation based on fast iterative
shrinkage-thresholding algorithm (FISTA) \cite{beck2009afast}
framework for this problem. This is because FISTA can be easily
applied for standard sparsity and joint sparsity, which could make
the validation of the benefit of the proposed model more convenient. In addition, the formulation (\ref{eqn:appro1}) can be easily extended to the
combination of total variation (TV) via the Fast
Composite Splitting Algorithms (FCSA) scheme
\cite{huang2011efficient}.
Note that other algorithms may be used to solve the forest sparsity problems, e.g. \cite{deng2011group,jacob2009group,kowalski2012social}, but determining the optimal algorithm for forest sparsity is not the scope of this article.

FISTA \cite{beck2009afast} is a accelerated version of proximal
method which minimizes the object function with the following form:
\begin{eqnarray}
\min \{F(x) = f(x) + g(x)\} \label{eqn:FISTA}
\end{eqnarray}
where $f(x)$ is a convex smooth function with Lipschitz constant
$L_f$ and $g(x)$ is a convex but usually nonsmooth function.  It comes to the original FISTA when $f(x) = \frac{1}{2}||Ax-b||_2^2$ and $g(x) =
\lambda \|\Phi x\|_{1}$, which is summarized in Algorithm \ref{alg:FISTA},
where, $A^T$ denotes the transpose of $A$.

\begin{figure}[h]
%\centering
\begin{algorithm}[H]
\caption{FISTA \cite{beck2009afast}} \label{alg:FISTA}
\begin{algorithmic}
\STATE {\bfseries Input:} $\rho={1}/{L_f}$, $\lambda$, $n=1$,
$t^{1}=1$ $r^1=x^{0}$
\WHILE{not meet the stopping criterion}{
\STATE $y=r^{n}-\rho A^T(Ar^{n}-b)$
\STATE $x=\arg\min_{x}\{\frac{1}{2\rho}\|x-y\|^2+\lambda\|\Phi x\|_{1}\}$
\STATE  $t^{n+1}={1+\sqrt{1+4(t^{n})^{2}}}/{2}$
\STATE  $r^{n+1}=x^{n}+\frac{t^{n}-1}{t^{n+1}}(x^{n}-x^{n-1})$
\STATE $n=n+1$
}
\ENDWHILE
\end{algorithmic}
\end{algorithm}
\end{figure}

For the second step, there is closed form solution by soft-thresholding. For joint sparsity problem where $g(x) = \lambda\|\Phi x\|_{2,1}$, the second step also has closed form solution. %For multi-channel data with joint sparsity, the $\ell_{2,1}$ norm denotes the summation of $\ell_{2}$ norm of the coefficients at the same position across channels.
We call the version as FISTA\_Joint for joint sparsity. However, for the problem (\ref{eqn:appro1}) with overlapped groups, we can not directly apply FISTA to solve it.

In order to transfer the problem (\ref{eqn:appro1}) to
non-overlapping version, we introduce a binary matrix $G \in
\mathbb{R}^{D\times TN}$ ($D>TN$) to duplicate the overlapped
coefficients. Each row of $G$ only contains one 1 and all else are
0s. The 1 appears in the $i$-th column corresponds to the $i$-th
coefficient of  $\Phi x$. Intuitively, if the coefficient is
included in $j$ groups, $G$ will contains $j$ such rows. An
auxiliary variable $z$ is used to constrain $G \Phi x$. This scheme
is widely utilized in the alternating direction method (ADM)
\cite{deng2011group}. The alternating formulation becomes:
%For
%problem (\ref{eqn:appro1}), if we let $f(x) = \frac{1}{2}||Ax-b||_2^2$ and $g(x) =
%\lambda \sum_{g \in \mathcal{G}} ||(G \Phi x)_g||_{2}$, there will be no closed form solution
%for (\ref{eqn:FISTAxk}). we introduce an auxiliary variable $z$
%constrain $G \Phi x$. The formulation then becomes to:
\begin{eqnarray}\vspace{-0.0cm}
\min_{x,z} \{\frac{1}{2}\|Ax-b\|_2^2+\lambda \sum_{g \in
\mathcal{G}} ||z_{g}||_2+ \frac{\gamma}{2} ||z-G\Phi x||_2^2 \}
\label{eq:uncPro}\vspace{-0.0cm}
\end{eqnarray}
where $\gamma$ is another positive parameter. We iteratively solve
this alternative formulation by minimizing $x$ and
$z$ subproblems respectively. For the $z$ subproblem:
\begin{eqnarray}
\hat{z_{g}}=\arg \min_{z_{g}}\{\lambda ||z_{g}||_2+ \frac{\gamma}{2}
||z_{g}-(G\Phi x)_{g}||_2^2 \}  ,g \in \mathcal{G} \label{eqn:zsub}
\end{eqnarray} \vspace{-0.0cm}
which has the closed form solution:
\begin{eqnarray}
\hat{z_{g}}= \max(||(G\Phi x)_{g}||_{2}-\frac{\lambda}{\gamma},0)\frac{(G\Phi x)_{g}}{||(G\Phi x)_{g}||_2} ,g
\in \mathcal{G}
 \label{eqn:zsolution}
\end{eqnarray} \vspace{-0.0cm}
We denote it as a shrinkgroup operation. For the $x$-subproblem:
\begin{eqnarray}
\hat{x} = \arg \min_x \{ \frac{1}{2}\|Ax-b\|_2^2+ \frac{\gamma}{2}
||z-G\Phi x||_2^2 \} \label{eqn:xsub}
\end{eqnarray} \vspace{-0.0cm}
The optimal solution is $x=(A^TA+\lambda \Phi^T G^T  G \Phi
)^{-1}(A^Tb+\lambda \Phi^T G^Tz)$, which contains a large scale inverse
problem. Actually, this problem can be efficient solved by various methods. In order to compare with FISTA and FISTA\_Joint, we apply FISTA to solve (\ref{eqn:xsub}). This could demonstrate the benefit of forest sparsity more clearly. Let $f(x)= \frac{1}{2}\|Ax-b\|_2^2 + \frac{\lambda}{2}
||z-G\Phi x||_2^2$ and $g(x)=0$. Supposing its Lipschitz constant to be $L_{f}$,
the whole algorithm is summarized in Algorithm \ref{alg:FISFO}.

\begin{figure}[htbp]
\centering
%\begin{minipage}{3in}
 \begin{algorithm}[H]
\caption{FISTA\_Forest} \label{alg:FISFO}
\begin{algorithmic}
   \STATE {\bfseries Input:} $\rho=1/L_{f}$, $r^{1}=x^{0}$, $t^{1}=1,\lambda, \gamma, n=1$
   %\REPEAT
   %\FOR{$k=1$ {\bfseries to} $N$}
   \WHILE{not meet the stopping criterion}{
   \STATE $z=shrinkgroup(G\Phi x^{n-1},\lambda/ \gamma)$
   \STATE $x^n=r^{n}-\rho [A^T(Ar^{n}-b)+\gamma \Phi^T G^T (G \Phi r^{n} -z)]$
   \STATE $ t^{n+1}=[1+\sqrt{1+4(t^{n})^{2}}]/2$
   \STATE $ r^{n+1}=x^{n}+\frac{t^{n}-1}{t^{n+1}}(x^{n}-x^{n-1})$
    \STATE $n=n+1$
 %  \ENDFOR
       }
   \ENDWHILE
   %\UNTIL{Stop criterions}
\end{algorithmic}
\end{algorithm}
%\end{minipage}
\end{figure}

For the first step, we solve (\ref{eqn:zsub}) while
$\frac{1}{2}\|Ax-b\|_2^2$ keeps the same. The object function value
in (\ref{eq:uncPro}) decreases. For the second step,
(\ref{eqn:xsub}) is solved by FISTA iteratively while $\lambda
\sum_{g \in\mathcal{G}} ||z_{g}||_2$ keeps the same. Therefore, the
object function value in (\ref{eq:uncPro}) decreases in each
iteration and the algorithm is convergent. Algorithm \ref{alg:FISFO}
is also used to implement tree sparsity by recovering the data
channel-by-channel separately. We call it FISTA\_Tree.

%where $\nabla f(r^{n}) = A^T(Ar^{n}-b)+\gamma \Phi^T G^T (G \Phi r^{n} -z)$.
In some practical applications, the data tends to be forest-sparse but not strictly. We can soften and complement the forest assumption with other penalties, such as joint $\ell_{2,1}$ norm or TV. For example, after combining TV, problem (\ref{eq:uncPro}) becomes:
\begin{eqnarray}\vspace{-0.0cm}
\min_{x,z} \{\frac{1}{2}\|Ax-b\|_2^2+\lambda \sum_{g \in
\mathcal{G}} ||z_{g}||_2  + \frac{\gamma}{2} ||z-G\Phi x||_2^2 \notag \\ +\mu ||x||_{TV} \}
\label{eq:TVfor}\vspace{-0.0cm}
\end{eqnarray}
where $||x||_{TV} = \sum_{i=1}^{TN}\sqrt{(\nabla_{1}x_{i})^2+(\nabla_{2}x_{i})^2}$; $\nabla_{1}$ and $\nabla_{2}$ denote the forward finite
difference operators on the first and second coordinates respectively; $\mu$ is a positive parameter.
Compared with Algorithm \ref{alg:FISFO}, we only need to set $g(x) = \mu ||x||_{TV}$ and the corresponding subproblem has already been solved \cite{beck2009afast,huang2011efficient,huang2011composite}. This TV combined algorithm is called FCSA\_Forest, which will be used in the experiments. To avoid repetition, it is not listed.

%To combine total variation regularization, we only need to let $g(x) = ||x||_{TV}$ for the $x$-subproblem.
%Note that this algorithm could solve the combined model with both forest sparsity and total variation.

%\section{Application} \label{sec:app}

%\subsection{Multi-contrast MRI}

%\subsection{pMRI}

%\subsection{Color Images}

%
%Each color channel can be
%represented by wavelet tree sparsity. If further exploited with
%forest sparse model, this result would be reasonably better.

%\subsection{Multispectral Images}

\section{Applications and Experiments} \label{sec:exp}

We conduct experiments on the RGB color image, multi-contrast MR images,
MR image of multi-channel coils and the multispectral image to validate
the benefit of forest sparsity. All experiments are conducted on a
desktop with 3.4GHz Intel core i7 3770 CPU. Matlab version is
7.8(2009a). %The matlab code to reproduce all the experimental results can be downloaded\footnote{\texttt{https://dl.dropbox.com/u/58080329/code\_TSP\_forest.zip}}.
%For MR images, the sampling matrix is a partial Fourier
%transform. For color image and multispectral images, the sampling
%matrix is a random projection matrix.
If the sampling matrix $A$ is $M$ by $N$, the sampling ratio is
defined as $M/N$. All measurements are mixed with Gaussian white
noise of 0.01 standard deviation. Signal-to-Noise Ratio (SNR) is
used as the metric for evaluations:
\begin{eqnarray}
SNR=10\log_{10}(V_{s}/V_{n})
\end{eqnarray}
where $V_{n}$ is the Mean Square
Error between the original data $x_{0}$ and the reconstructed
$x$; $V_{s}=var(x_{0})$ denotes the power level of the original
data where $var(x_{0})$ denotes the variance of the values in
$x_{0}$.

%The reproducible MATLAB code for the experiments in this paper is included into the supplementary
%material. More results can be found there.

\subsection{Multi-contrast MRI}

Multi-contrast MRI is a popular technique to aid clinical diagnosis. For example T1
weighted MR images could distinguish fat from water, with water
appearing darker and fat brighter. In T2 weighted images fat
is darker and water is lighter, which is better suited to imaging
edema. Although with different intensities, T1/T2 or proton-density
weighted MR images are scanned at the same anatomical position.
Therefore they are not independent but highly correlated. Multi-contrast MR images are typically forest-sparse under the wavelet
basis. Suppose
$\{x_{t}\}_{t=1}^{T} \in \mathbb{R}^{N}$ are the multi-contrast
images for the same anatomical cross section and
$\{b_{t}\}_{t=1}^{T}$ are the corresponding
undersampled data in Fourier domain, the forest-sparse
reconstruction can be formulated as:
\begin{eqnarray}
\hat{x}=\arg\min_{x}\|\Phi x\|_{\mathcal{F},T} + \lambda
\sum_{s=1}^{T}\|R_{t}x_t-b_{t}\|^2\label{eqn:spgl1GS}
\end{eqnarray}
where $x$ is the vertorized data of $[x_{1}, ..., x_{T}]$ and $R_{t}$ is the measurement matrix
for the image $x_{t}$. This is an extension of conventional CS-MRI \cite{lustig2007sparse}.
Fig. \ref{fig:mtforest} shows an example of
the forest structure in multi-contrast MR images.

\vspace{-0.0cm}
\begin{figure}[htbp]
\centering \vspace{-0.0cm}
        \includegraphics[scale=0.18]{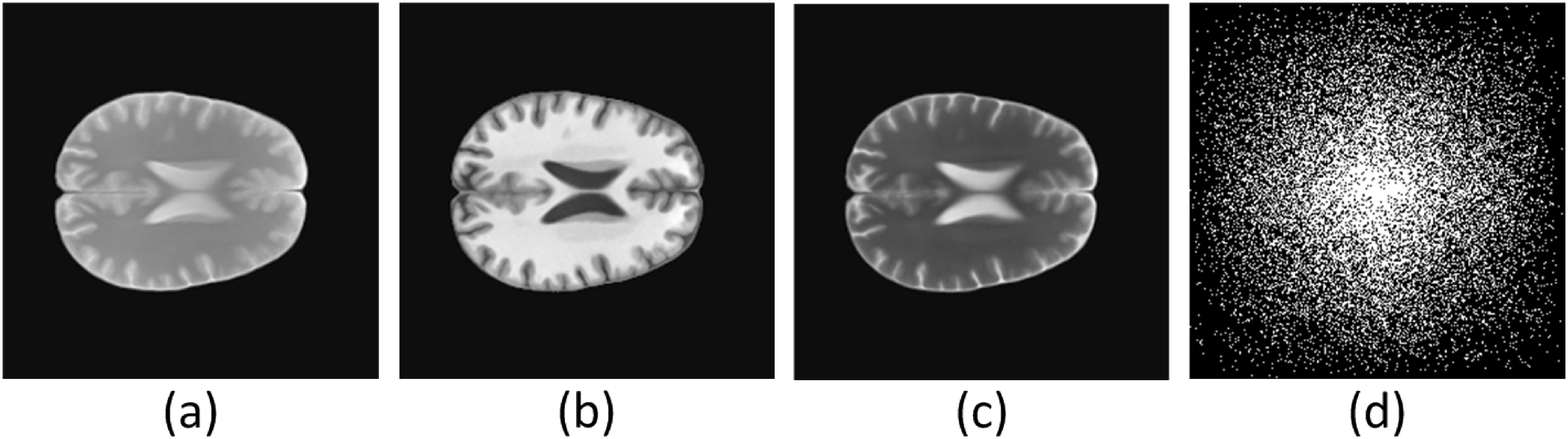}
\caption{(a)-(c) the original multi-constrast images  and (d) the sampling mask.}\label{fig:mtori}\vspace{-0.0cm}
\end{figure} \vspace{-0.0cm}

The data is extracted from the SRI24 Multi-Channel Brain
Atlas Dataset \cite{rohlfing2009sri24}. In the Fourier domain, we randomly obtain more samples
in low frequencies and less samples in higher frequencies.
This sampling scheme has been widely used for CS-MRI \cite{lustig2007sparse,ma2008efficient,huang2011efficient}. Fig. \ref{fig:mtori} shows the original multi-contrast MR images and the sampling mask.

\begin{figure}[htbp]
\centering \vspace{-0.0cm}
   \subfigure[]{\label{fig:ST_TimeSNR}
        \includegraphics[scale=0.4]{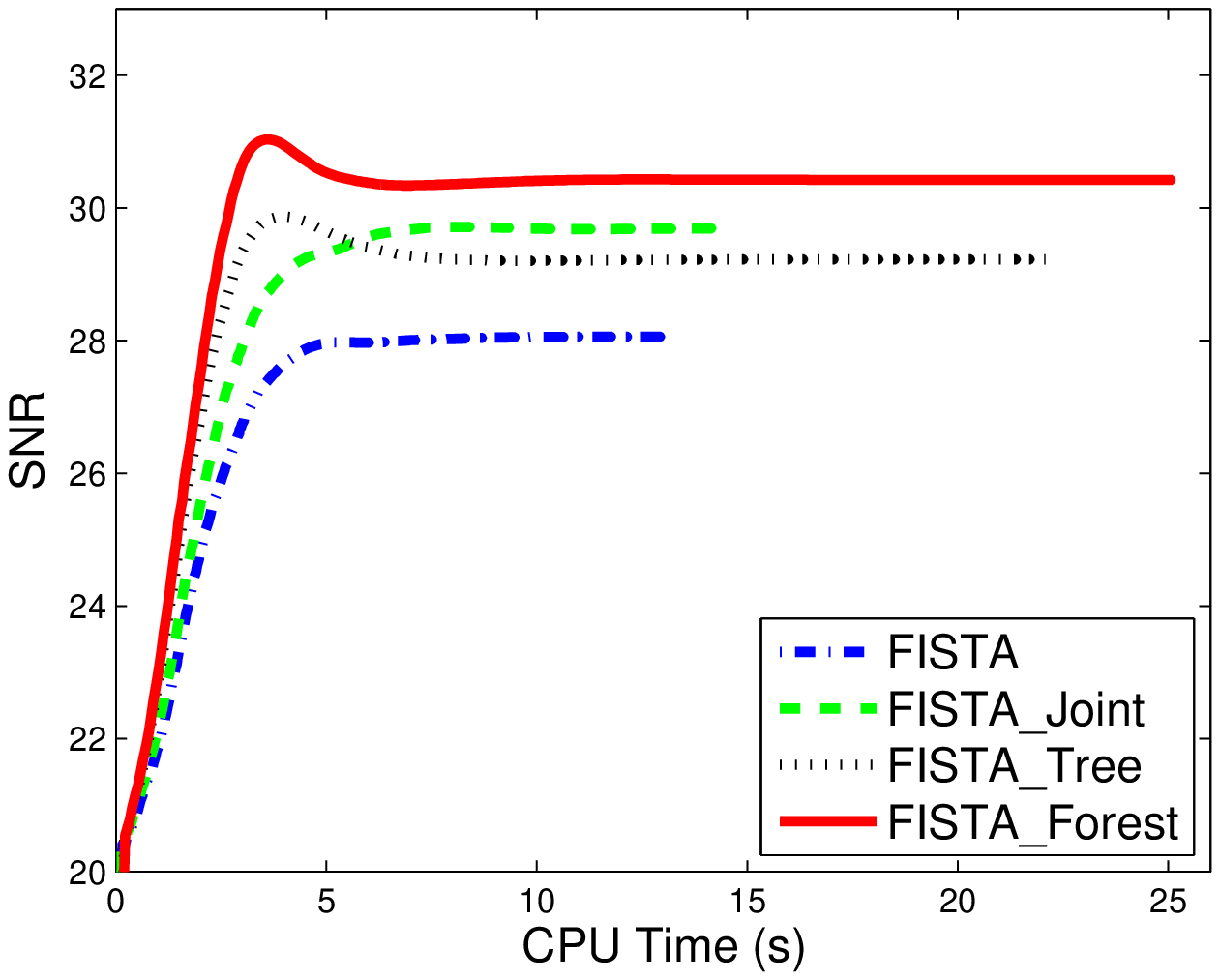}}
    \subfigure[]{\label{fig:MT_TimeSNR}
        \includegraphics[scale=0.4]{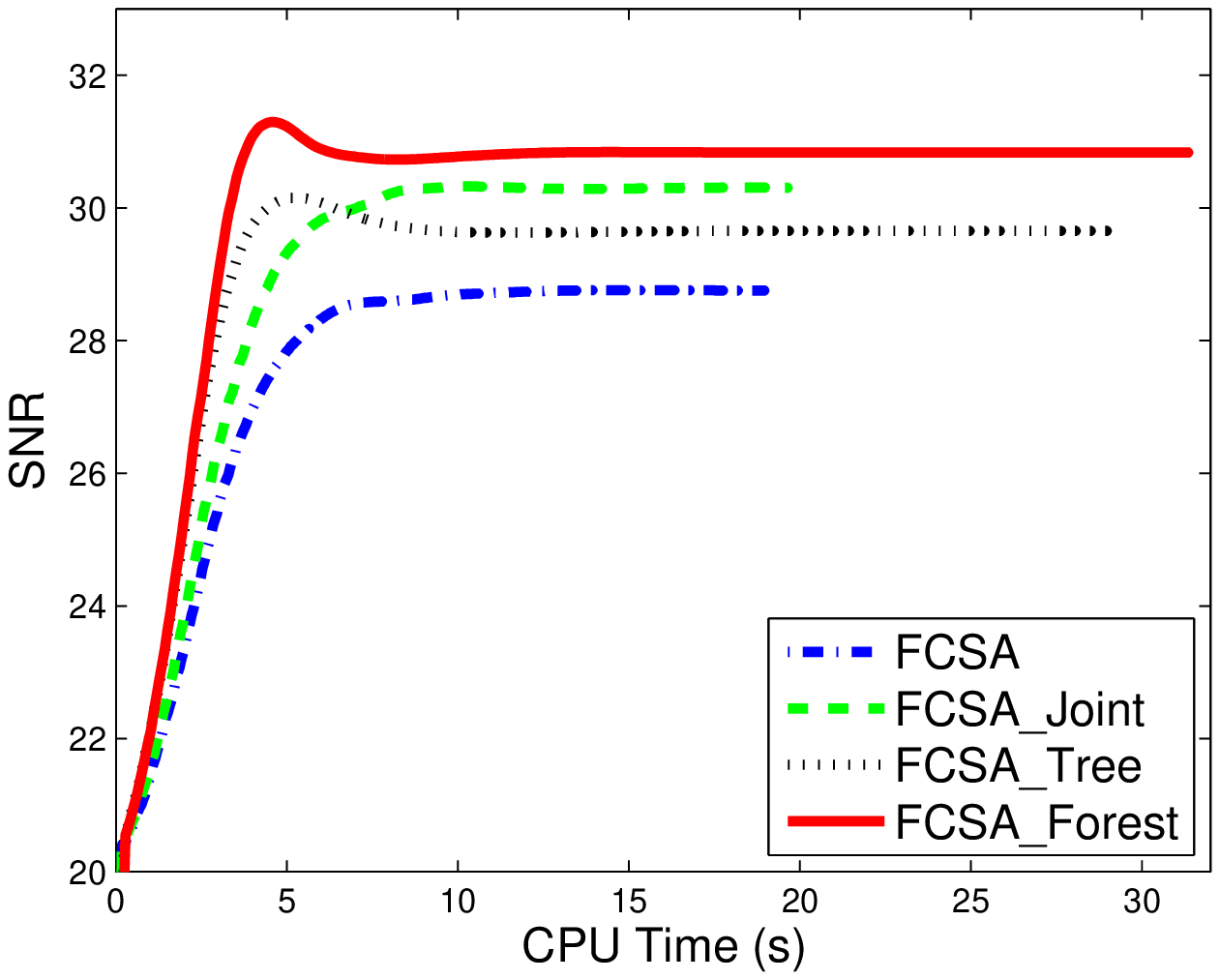}}
\caption{Performance comparisons among different algorithms. (a) Multi-constrast MR images reconstruction with $20\%$ sampling. Their final SNRs are 28.05, 29.69, 29.22 and 30.42 respectively. The time costs are 13.11s, 14.43s, 22.08s and 25.11s respectively. (b) Multi-constrast MR images reconstruction with $20\%$ sampling by both wavelet sparsity and TV regularization. Their final SNRs are 28.75,   30.30, 29.65 and 30.83 respectively. The time costs are 19.00s, 19.68s, 29.11s and 31.41s respectively.}\label{fig:MTMRI}\vspace{-0.0cm}
\end{figure}

We compare four algorithms on this dataset: FISTA, FISTA\_Joint, FISTA\_Tree and FISTA\_Forest.  The parameter $\lambda$ is set 0.035, and $\gamma$ is set to $0.5\lambda$. We run each algorithm 400 iterations. Fig. \ref{fig:MTMRI} (a) demonstrates the performance comparisons among different algorithms. From the figure, we could observe that
modeling with forest sparsity achieves the highest SNR after convergence. Although the algorithm for forest sparsity takes more time due to the overlapping structure, it always outperforms all others in terms of accuracy.

In addition, as total variation is very popular in CS-MRI \cite{lustig2007sparse,huang2011efficient,huang2012fast}, we compare our FCSA\_Forest algorithm with FCSA \cite{huang2011efficient} (TV is combined in FISTA), FCSA\_Joint \cite{huang2012fast} (TV is combined in FISTA\_Joint) and FCSA\_Tree.
% it in all these algorithms via the FCSA \cite{huang2011efficient}\cite{huang2012fast} scheme.
%we combine total variation regularization in all algorithms and then all the algorithms become in the FCSA \cite{huang2011efficient}\cite{huang2012fast} scheme.
The parameter $\mu$ for TV is set $0.001$, the same as that in previous works \cite{ma2008efficient,huang2011efficient}. Fig. \ref{fig:MTMRI} (b) demonstrates the performance comparison including TV regularization. Compared with  Fig. \ref{fig:MTMRI} (a), all algorithms improve at different degrees. However, the ranking does not change, which validates the superiority of forest sparsity.  As FCSA has been proved to be better than other algorithms for general compressive sensing MRI (CS-MRI) \cite{lustig2007sparse,ma2008efficient,yang2010fast} and FCSA\_Joint \cite{huang2012fast} better than \cite{bilgic2011multi,majumdar2011joint} in multi-contrast MRI, the proposed method further improves CS-MRI and make it more feasible than before.

%\vspace{-0.0cm}
%\begin{figure*}[htbp]
%\centering \vspace{-0.0cm}
%        \includegraphics[scale=0.23]{figures/MTSNR.jpg}
%\caption{Performance comparisons among different algorithms. (a) Multi-constrast MR images reconstruction with $20\%$ sampling. Their final SNRs are 27.87,    29.37, 29.05 and 30.14 respectively. The time costs are 13.81s, 14.87s, 23.70s and  25.41s respectively. (b) Multi-constrast MR images reconstruction with $25\%$ sampling. Their final SNRs are 30.30, 32.49, 31.57 and    33.01 respectively. The time costs are 13.89s, 14.91s, 22.40s and 25.38s respectively. (c) Multi-constrast MR images reconstruction with $20\%$ sampling by both wavelet sparsity and TV regularization. Their final SNRs are 28.56,    29.96, 29.46 and 30.54 respectively. The time costs are 19.69s, 19.57s, 29.30s and 31.12s respectively.}\label{fig:MTMRI}\vspace{-0.0cm}
%\end{figure*} \vspace{-0.0cm}

\vspace{-0.0cm}
 \vspace{-0.0cm}

\begin{figure}[htbp]
\centering \vspace{-0.0cm}
        \includegraphics[scale=0.5]{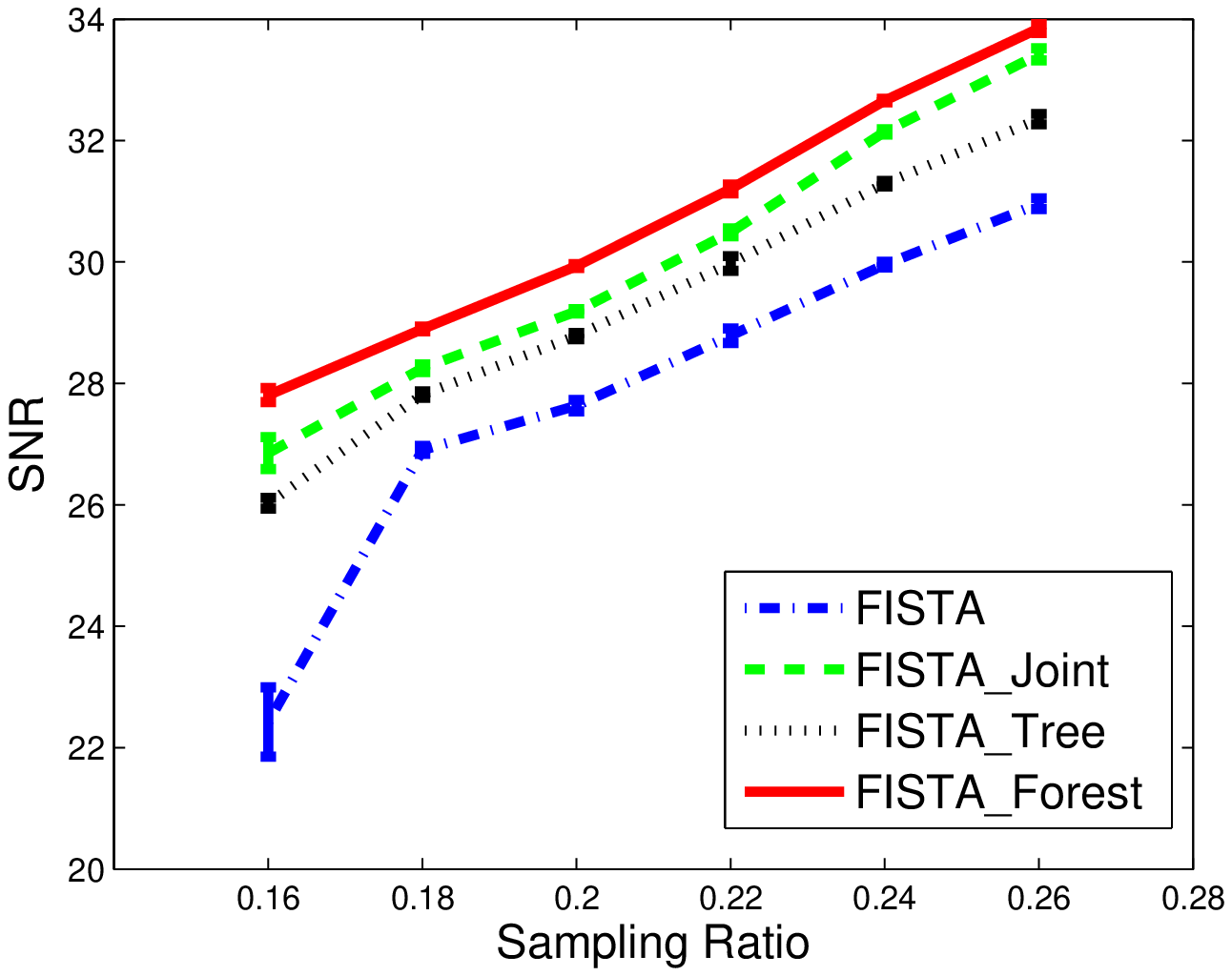}
\caption{Reconstruction performance with different sampling ratios.}\label{fig:ratios}\vspace{-0.0cm}
\end{figure}

In order to validate the benefit of forest sparsity in terms of
measurement number, we conduct an experiment to reconstruct
multi-contrast MR images from different sampling ratios. Fig.
\ref{fig:ratios} demonstrates the final results of four algorithms
with sampling ratio from $16\%$ to $26\%$. With more sampling, all
algorithms have better performance. However, The forest sparsity
algorithm always achieves the best reconstruction. For the same
reconstruction accuracy, the FISTA\_Forest algorithm only requires
about $16\%$ measurements to achieve SNR 28, which is approximately
$2\%$, $3\%$, $5\%$ less than that of FISTA\_Joint, FISTA\_Tree and
FISTA respectively. More results of forest sparsity on multi-contrast MRI can be found in \cite{chen2014exploiting}.

\vspace{-0.0cm}
 \vspace{-0.0cm}

\subsection{parallel MRI}
To improve the scanning speed of MRI, an efficient and feasible way
is to acquire the data in parallel with multi-channel coils. The
scanning time depends on the number of measurements in Fourier
domain, and it will be significantly reduced when each coil only
acquires a small fraction of the whole measurements. The bottleneck is
how to reconstruct the original MR image efficiently and precisely.
This issue is called pMRI in literature. Sparsity techniques have
been used to improve the classical method SENSE \cite{pruessmann1999sense}.
However, when the coil sensitivity can not be estimated precisely,
the final image would contain visual artifacts. Unlike previous
CS-SENSE \cite{liang2009accelerating} which reconstructs the images of multi-coils
individually, calibrationless parallel MRI \cite{majumdar2012calibration,chen2013calibrationless} recovers the aliased
images of all coils jointly by assuming the data is jointly sparse.

\begin{figure}[htbp]
\centering \vspace{-0.0cm}
    \subfigure[]{\label{fig:brain}
        \includegraphics[scale=0.15]{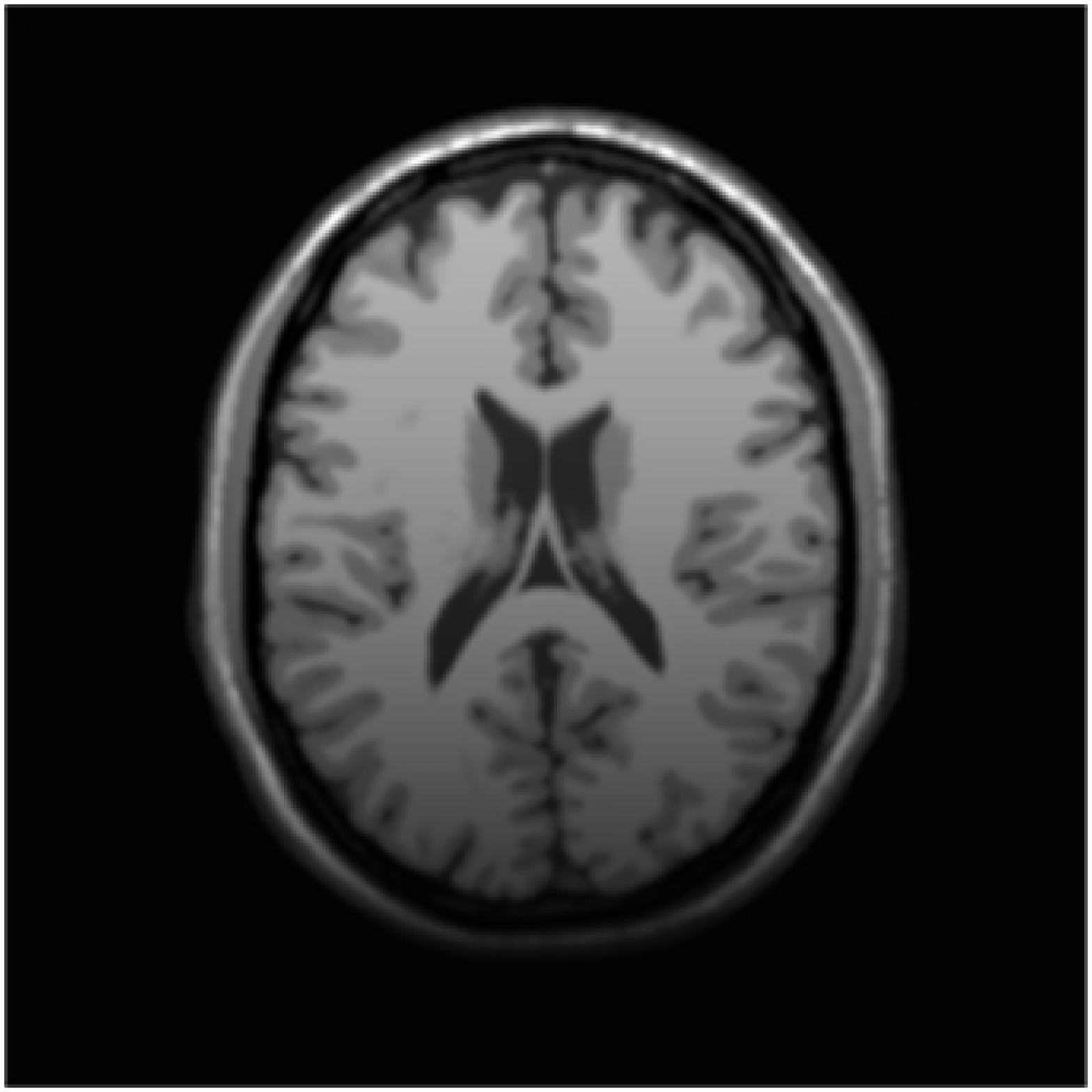}}
    \subfigure[]{\label{fig:tree}
        \includegraphics[scale=0.15]{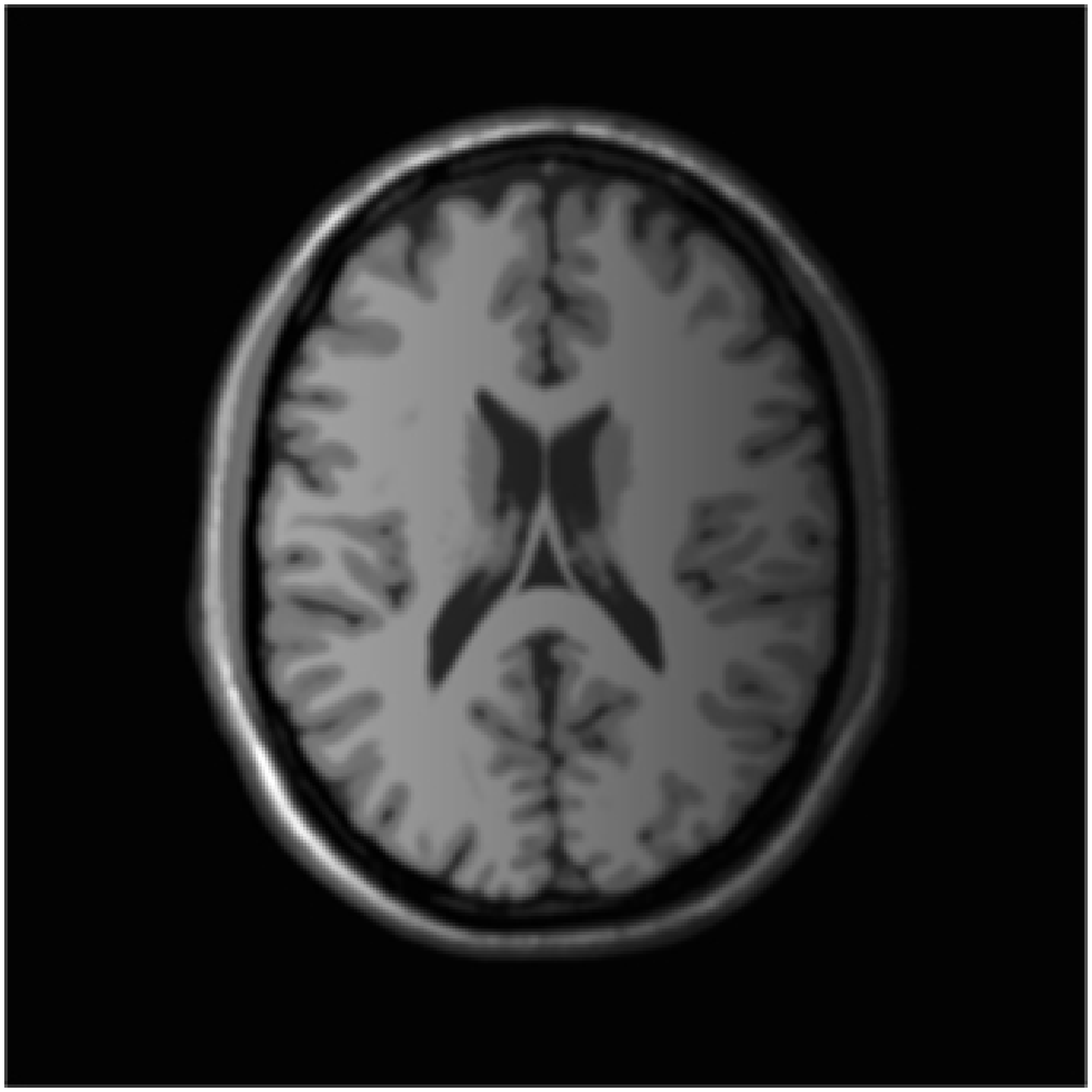}}
    \subfigure[]{\label{fig:brain}
        \includegraphics[scale=0.15]{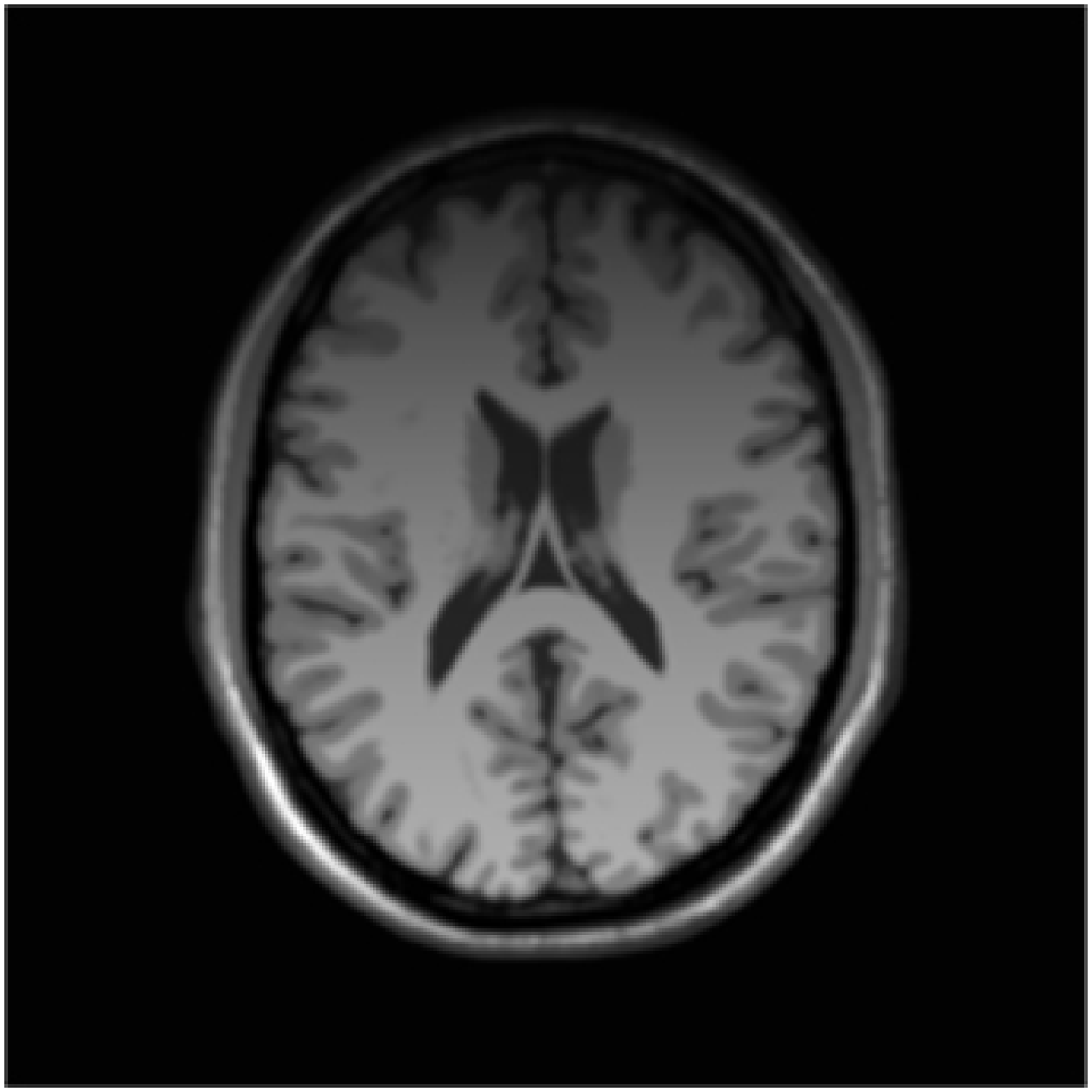}}
    \subfigure[]{\label{fig:tree}
        \includegraphics[scale=0.15]{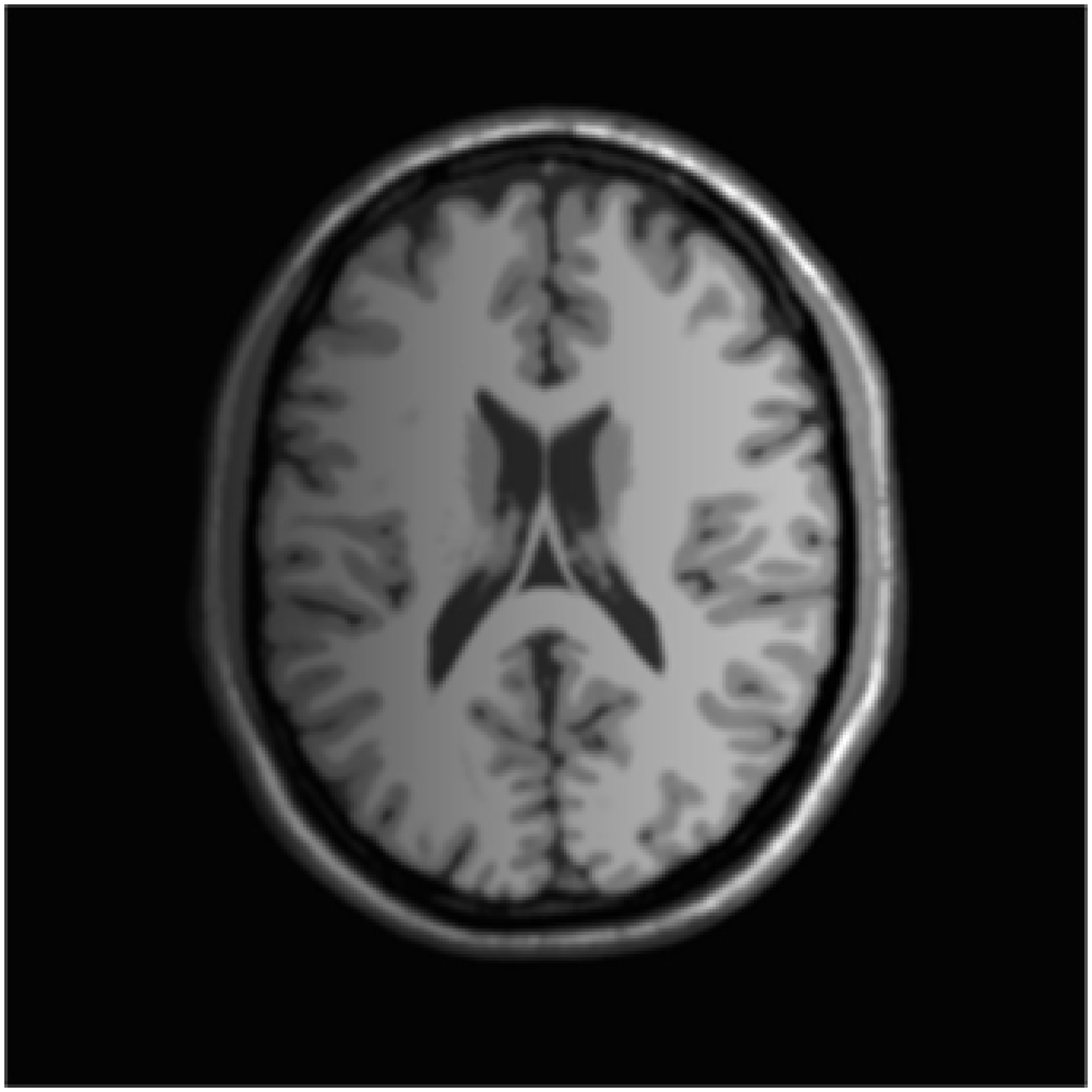}}
\caption{The aliased MR images of multi-coils. Due to the different locations of the coils, they have different sensitivities to the same image.}\label{fig:pmri}\vspace{-0.0cm}
\end{figure}

Let $T$ equal to the number of coils and $b_t$ be the measurement vector from coil $t$. It is therefore the same CS problem as (\ref{eqn:spgl1GS}). The final result of CaLM-MRI is obtained by a sum of square (SoS) approach
without coil sensitivity and SENSE encoding. It shows comparable
results with those methods which need precise coil configuration.
As shown in Fig. \ref{fig:pmri}, the appearances of different images obtained from multi-coils are very similar.
This method can be improved with forest sparsity, since the images follow the forest sparsity assumption.

\vspace{-0.0cm}
 \vspace{-0.0cm}

There are two steps for compressive sensing pMRI reconstruction in  CaLM-MRI
\cite{majumdar2012calibration}: 1) the aliased images are recovered from the undersampled Fourier signals of different coil channels by CS methods; 2) The final image for clinical diagnosis is synthesized by the recovered aliased images using the sum-of-square (SoS) approach. As discussed above, these aliased images should be forest-sparse under the wavelet basis.
We compare our algorithm with FISTA\_Joint and SPGL1 \cite{van2008probing} which solves the joint $\ell_{2,1}$ norm problem in CaLM-MRI. For the second step, all methods use the SoS approach from the aliased images that they recovered. All algorithms run enough time until it has converged.

\begin{table}[h!]
\caption{Comparisons of SNRs (\textnormal{d}B) on different sampling ratios with $4$ coils}
\label{ratio}
\begin{center}
%\newsavebox{\tbox}
\begin{lrbox}{\tbox}
\begin{tabular}{c|c|cccc}
\hline
%& \multicolumn{1}{c}{sampling ratios} &\multicolumn{1}{c}{25\%}
%&\multicolumn{1}{c}{20\%} &\multicolumn{1}{c}{17\%}
%&\multicolumn{1}{c}{15\%} \\
 & sampling ratios & 25\% & 20\% & 17\% & 15\%\\

\hline
SNR of & SPGL1      &26.72  &24.59 &23.08  &22.31 \\
Aliased Images & FISTA\_Joint   &26.95  &24.73  &23.06  &22.21\\
&FISTA\_Forest     &\textbf{27.47}  &\textbf{25.22} &\textbf{23.37}  &\textbf{22.59} \\
\hline
SNR of &SPGL1    &20.64 &20.35  &19.12  &18.64 \\
Final Image& FISTA\_Joint    &20.79 &20.41  &19.75  &18.49\\
&FISTA\_Forest   &\textbf{22.62}  &\textbf{22.29} &\textbf{21.03}  &\textbf{20.47} \\
\hline
\end{tabular}
\end{lrbox}
\scalebox{1}{\usebox{\tbox}}
\end{center}
\end{table}
%\vspace{-0.5cm}

\begin{table}[h!]
\caption{Comparisons of SNRs (\textnormal{d}B) on different number of coils with $20\%$ sampling}
\label{coils}
\begin{center}
%\newsavebox{\tbox}
\begin{lrbox}{\tbox}
\begin{tabular}{c|c|cccc}
\hline
%& \multicolumn{1}{c}{number of coils} &\multicolumn{1}{c}{2}
%&\multicolumn{1}{c}{4} &\multicolumn{1}{c}{6}
%&\multicolumn{1}{c}{8} \\
 & number of coils & 2 & 4 & 6 & 8\\

\hline
SNR of &  SPGL1    &23.33   &24.61  &24.74  &25.16 \\
Aliased Images &FISTA\_Joint   &23.41   &24.71  &24.89  &25.23 \\
& FISTA\_Forest    &\textbf{24.25}  &\textbf{25.12} &\textbf{25.29} &\textbf{25.52} \\
\hline
SNR of & SPGL1      &21.76  &18.95  &21.05  &21.32 \\
Final Image &FISTA\_Joint  &21.90 &18.94    &21.15  &21.87 \\
& FISTA\_Forest   &\textbf{22.44}   &\textbf{22.22} &\textbf{22.52} &\textbf{22.52} \\
\hline
\end{tabular}
\end{lrbox}
\scalebox{1}{\usebox{\tbox}}
\end{center}
\end{table}

Table \ref{ratio}  and Table \ref{coils}  show all the comprehensive comparisons among these algorithms. %The SNR for step 1 is calculated by concatenating all the recovered images to a vector, while the SNR for step 2 is the evaluation after the SoS for these images.
%of images obtained in the first step and the second step on various
%sampling ratios, and various numbers of coils for pMRI simulation.
For the same algorithm, more measurements or more number of coils tend
to increase the SNRs of aliased images, although it does not result in linear
improvement for the final image reconstruction. Another
observation is that FISTA\_Joint and SPGL1 have similar performance in terms of SNR on this data. This is because both of them solve the same joint sparsity problem, even with different schemes. Upgrading the model to forest sparsity, significant improvement can be gained. Finally, it is unknown how to combine TV in SPGL1. However, both FISTA\_Joint and FISTA\_Forest can easily combine TV, which can further enhance the results \cite{huang2012fast}.

%Another
%observation is that the model of forest sparsity is always better
%than joint sparsity due to the tree structure is utilized. The data is not only assumed to be joint sparse but "joint" tree sparse.
%More importantly , it is very convenient to combine total variation
%penalty in the proposed algorithm, but
%unknown how to do it in SPGL1.

%\subsection{Multispectral images Reconstruction}

\subsection{Color Image Reconstruction}

Color images captured by optical camera can be represented as
combinations of red, green, blue three colors. Different colors
synthesized by these three colors seems realistic to human eyes. By
observing the color channels are highly correlated, joint sparsity prior is
utilized in recent recovery \cite{majumdar2010compressive}. Modeling  with
$\ell_{2,1}$ norm regularization can gain additional SNR to standard
$\ell_{1}$ norm regularization. Further more, each color channel tends to be wavelet tree-sparse. If we model the problem with forest sparsity, this result would be reasonably better.

\vspace{-0.0cm}
\begin{figure}[htbp]
\centering \vspace{-0.0cm}
        \includegraphics[scale=0.4]{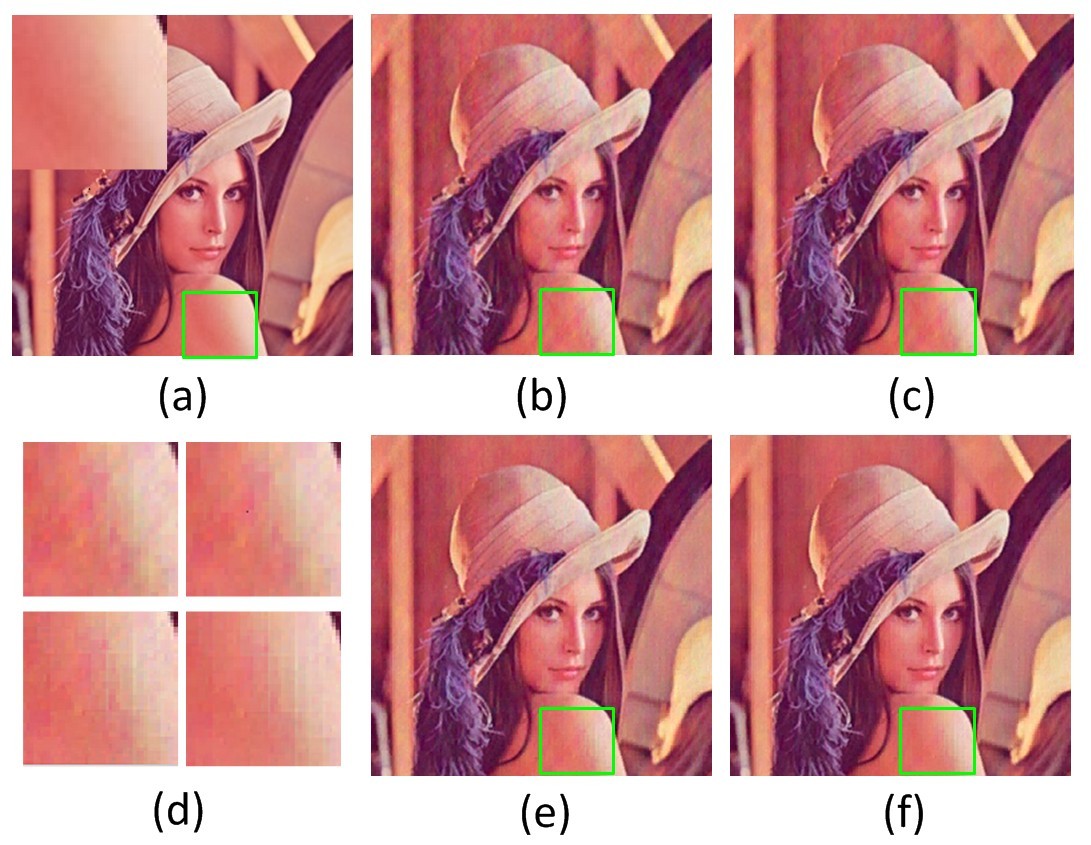}
\caption{Visual comparisons on the lena image reconstruction after
50 iterations with about $20\%$ sampling. (a) the original image and the patch detail; (b) recovered by FISTA; (c) recovered by
FISTA\_Joint; (d) the patch details for each recovered image; (e) recovered by FISTA\_Tree; (f) recovered by FISTA\_Forest. Their SNRs
are 16.65, 17.41, 17.66 and 18.92, respectively.}
\label{fig:fgbrec}\vspace{-0.0cm}
\end{figure} \vspace{-0.0cm}
%As mentioned in the last section, the RGB channels of color images should share a common support as they are highly correlated.
For color images, we compare our algorithm with FISTA, FISTA\_Joint and FISTA\_Tree.
Fig. \ref{fig:fgbrec} shows the visual results recovered by different sparse penalties.
Only after 50 iterations, the image recovered by our algorithm is very close to the original
one with the fewest artifacts (shown in the zoomed region of interest).

\subsection{Multispectral Image Reconstruction}

\vspace{-0.0cm}
\begin{figure}[h!]
\centering \vspace{-0.0cm}
        \includegraphics[scale=0.40]{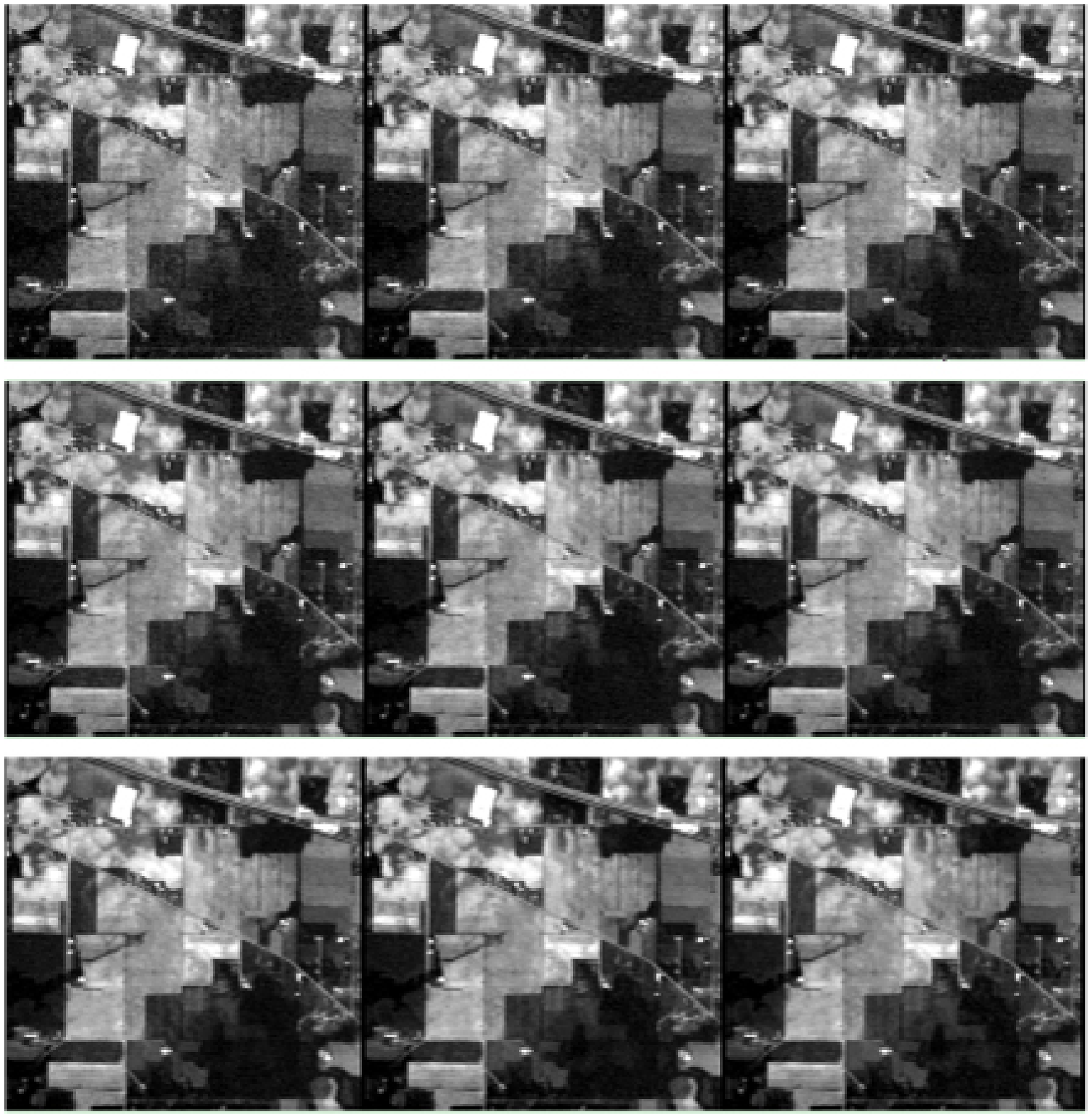}
\caption{The original multispectral image: band 6 to band 14. }\label{fig:mts}\vspace{-0.0cm}
\end{figure} \vspace{-0.0cm}

Different from common color images, a multispectral or hyperspectral
image is consisted of much more bands, which provides both spatial
and spectral representations of scenes. It is widely utilized on remote
sensing with applications to agriculture, environment detection etc..
However, the collection of large amount of data costs both
huge imaging time and storage space. By compressive sensing data
acquisition, the cost of imaging for remote sensing data could be
significantly reduced \cite{ma2009single}. Like RGB images, the bands of
multispectral image should represent the same scene. Each band has
tree sparsity property. Therefore, they follow the forest sparsity assumption.
Fig. \ref{fig:mts} shows bands 6 to 14 of a multispectral image of 1992 AVIRIS
Indian Pine Test Site 3 \footnote{The data is downloaded from
\texttt{https://engineering.purdue.edu \\ /$\scriptsize{\sim}$biehl/MultiSpec/hyperspectral.html}}.

\vspace{-0.0cm}
\begin{figure}[h!]
\centering \vspace{-0.0cm}
        \includegraphics[scale=0.5]{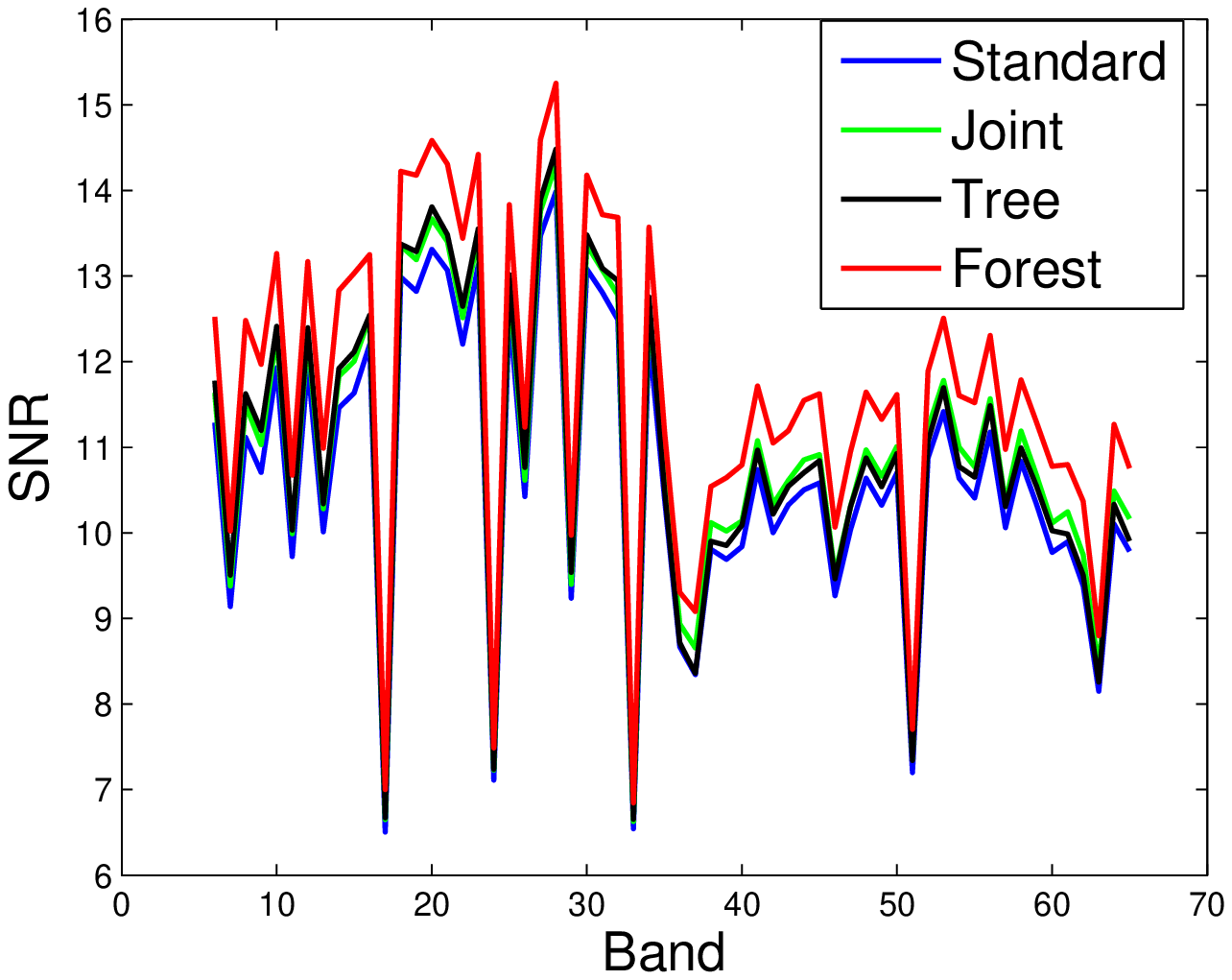}
\caption{Multispectral image reconstruction results by different
sparse models with about $20\%$ sampling.}
\label{fig:multiS}\vspace{-0.0cm}
\end{figure} \vspace{-0.0cm}

For multispectral image, we test a dataset of 1992 AVIRIS image
Indian Pine Test Site 3 (examples shown in Fig. \ref{fig:mts}). It is a $2 \times 2$ mile portion of Northwest Tippecanoe County of
Indiana. There are total $220$ bands. Each band is recovered separately for standard sparsity and tree sparsity, while every 3 bands are
reconstructed simultaneously by joint-sparse model and forest-sparse
model. Each image is cropped to $128\times 128$ for convenience. The number of
wavelet decomposition levels is set to 3. The SNRs of all recovered
images for band 6 to 66 are shown in Fig. \ref{fig:multiS}. One could observe that
modeling with forest sparsity always achieves the highest SNRs, which validates the benefit of forest sparsity.
%Experimental results on more bands could be found in the
%supplementary material.

%\subsection{Discussion}

\section{Conclusion} \label{sec:con}

In this paper, we have proposed a novel model \emph{forest sparsity} for
sparse learning and compressive sensing. This model enriches the family of structured sparsity and can be widely applied on numerous fields of sparse regularization problems. The benefit
of the proposed model has been theoretically proved and empirically validated in practical applications. Under compressive sensing
assumptions, significant reduction of measurements is achieved with
forest sparsity compared with standard sparsity, joint sparsity or
independent tree sparsity. A fast algorithm is developed to
efficiently solve the forest sparsity problem. While applying it on
practical applications such as multi-contrast MRI, pMRI, multispectral
image and color image reconstruction, extensive experiments
demonstrate the superiority of forest sparsity over standard
sparsity, joint sparsity and tree sparsity in terms of both accuracy
and computational complexity.

\appendices
%\section{Proof of the First Zonklar Equation}
%Appendix one text goes here.
%
%% you can choose not to have a title for an appendix
%% if you want by leaving the argument blank
%\section{}
%Appendix two text goes here.
\section{Proof of Theorem 2}
The proof is conducted on the binary tree case for convenience. The bound for quadtree can be easily extended.

First, we need to figure out the number of subtrees (size $k$) of a
binary tree (size $N$). Note that the root of the subtrees should be
the binary tree's root.

Case 1: when $k \leq \left \lfloor \log_2 N \right \rfloor$, the
number of subtrees of size $k$ is just the Catalan number:
\begin{eqnarray}
L_{\mathcal{T}} = \frac{1}{k+1}\binom{2k}{k} \leq \frac{(2e)^k}{k+1}
\leq \frac{e^k N}{k+1}.
\end{eqnarray} \vspace{-0.0cm}

Case 2: when $k>\left \lfloor \log_2 N \right \rfloor$, the number
of subtrees of size $k$ should follow \cite{baraniuk2010model}:
\begin{align}
L_{\mathcal{T}} & \leq \frac{4^k}{k}(\frac{6}{\sqrt{\pi
k}}\ln\frac{\log_2 N}{\left \lfloor \log_2 k \right \rfloor)} +
\frac{128}{e^2 \left \lfloor \log_2 k \right \rfloor} ) \notag \\
& \leq \frac{4^k}{k}( \frac{ c_1 \log_2 N}{\left \lfloor \log_2 k
\right
\rfloor)} + \frac{c2}{\left \lfloor \log_2 k \right \rfloor} ) \notag \\
&\leq \frac{4^k}{k} \frac{c_1 \log_2 (c_3 N )}{\log_2 k} \notag \\
&\leq \frac{4^k (c_4 N)}{k} .
\end{align} \vspace{-0.0cm}
where $c_1$, $c_2$, $c_3$, $c_4$ are some constants. Therefore we have:
\begin{eqnarray}
L_{\mathcal{T}} \leq \left\{\begin{matrix}
\frac{e^k N}{k+1} & \textrm{if} \ k \leq \left \lfloor \log_2 N \right \rfloor ,\\
\frac{4^k (c_4 N)}{k} & \textrm{if} \ k>\left \lfloor \log_2 N \right
\rfloor .
\end{matrix}\right. \label{eqn:Ltree}
\end{eqnarray} \vspace{-0.0cm}

According to Theorem 1:
\begin{eqnarray}
M \geq \frac{2}{c\delta}(\ln (2L) + k\ln \frac{12}{\delta}+t).
\label{eqn:L}
\end{eqnarray} \vspace{-0.0cm}
With (\ref{eqn:L}), the number of measurements should satisfy:
\begin{eqnarray}
M \geq \left\{\begin{matrix}
\ \frac{2}{c\delta_{\mathcal{T}_{k}}} ( k + \ln(N/(k+1)) + k\ln(12/\delta_{\mathcal{T}_{k}})  \\ + \ln 2+t) \qquad \textrm{if} \ k \leq \left \lfloor \log_2 N \right \rfloor , \\
\ \frac{2}{c\delta_{\mathcal{T}_{k}}} (k\ln4 + \ln(c_4N/k) + k\ln(12/\delta_{\mathcal{T}_{k}})
 \\ + \ln 2+t) \qquad \textrm{if} \ k>\left \lfloor \log_2 N \right \rfloor .
\end{matrix}\right.
\end{eqnarray} \vspace{-0.0cm}

For both cases, we have $ M = \mathcal{O}(k+\log(N/k))$ as the minimum number of measurements. Similar bound also has been proved in previous papers \cite{baraniuk2010model} \cite{huang2011learning}.

%\section{Proof of Lemma 2}
%
%From Lemma 1, the number of subtrees for a $k$-sparse tree is given in
%(\ref{eqn:Ltree}). Suppose there are $T$ trees in multichannel data and they are
%independent, the number of combinations should be
%$(L_{\mathcal{T}})^T$ and the sparsity number now is $Tk$. Therefore, the number of measurements have to
%satisfy:
%\begin{eqnarray}
%M \geq \left\{\begin{matrix}
%\ \frac{2}{c\delta_{\mathcal{T}_{T,k}}} ( Tk + T\ln(N/(k+1)) + Tk\ln(12/\delta_{\mathcal{T}_{T,k}})  \\ + \ln 2+t) \qquad \textrm{if} \ k \leq \left \lfloor \log_2 N \right \rfloor , \\
%\ \frac{2}{c\delta_{\mathcal{T}_{T,k}}} ( Tk\ln4 + T\ln(c_4N/k) + Tk\ln(12/\delta_{\mathcal{T}_{T,k}})
% \\ + \ln2  +t) \qquad \textrm{if} \ k>\left \lfloor \log_2 N \right \rfloor .
%\end{matrix}\right.  \notag
%\end{eqnarray} \vspace{-0.0cm}
%
%For both cases, the required bound is  $ M =
%\mathcal{O}(Tk+T\log(N/k))$. %This also can be obtained by $T \times
%%\mathcal{O}(k+\log(N/k)) = \mathcal{O}(Tk+T\log(N/k))$.

\section{Proof of Theorem 3}

If the data is forest-sparse, the support set of different trees are
dependent. It means if the support set for one tree is fixed, then all support sets for other trees are fixed. Accordingly, the number of combinations is still
$L_{\mathcal{T}}$. Note that the sparsity number is $Tk$ as there are $T$ trees. Therefore,
\begin{eqnarray}
TM \geq \left\{\begin{matrix}
\ \frac{2}{c\delta_{\mathcal{F}_{T,k}}} (  k + \ln(N/(k+1))
+ Tk\ln(12/\delta_{\mathcal{F}_{T,k}})  \\ + \ln 2 +t)  \qquad \textrm{if} \ k \leq \left \lfloor \log_2 N
\right \rfloor ,\\
\ \frac{2}{c\delta_{\mathcal{F}_{T,k}}} ( k\ln4 + \ln((c_4N)/k)
+ Tk\ln(12/\delta_{\mathcal{F}_{T,k}}) \\ + \ln2 +t)
  \qquad \textrm{if} \ k>\left \lfloor \log_2 N \right \rfloor .
\end{matrix}\right.
\end{eqnarray}

For both cases, the bound is reduced to $ TM =
\mathcal{O}(Tk+\log(N/k))$.

\section{Proof of Theorem 4}

We first derive the sufficient condition that guarantees the RIP for block-diagonal matrices.

%\textbf{Theorem : (DB $\mathcal{A}$-RIP)}  For
%any $t>0$, let
%\begin{eqnarray}
%TM \geq \frac{2T}{c_1 \min (c_2^2 \Gamma_2, c_2 \Gamma_\infty)}(\ln (2L) + k\ln \frac{12}{\delta_{\mathcal{A}_k}}+t) \label{eqn:mbound}
%\end{eqnarray}
%where $c_1$ and $c_2$ are absolute constants, $\Gamma_2 = \frac{(\sum_{t=1}^T ||x_t||^2_2 )^2 }{\sum_{t=1}^T ||x_t||^4_2}$ and $\Gamma_\infty = \frac{\sum_{t=1}^T ||x_t||^2_2  }{\max_{t=1}^T ||x_t||^2_2}$. There exist a matrix $A \in \mathbb{R}^{TM\times TN} $ composed by sub-Gaussian random matrices $\{A'_t \in \mathbb{R}^{M\times N} \}_{t=1}^T$ as in (\ref{eqn:bigA}) such
%\begin{eqnarray}
%(1-\delta_{\mathcal{A}})||X||_2^2 \leq ||AX||_2^2 \leq (1+\delta_{\mathcal{A}})||X||_2^2. \notag
%\end{eqnarray}
%holds for all $X=[x_1;x_2;...;x_T]\in \mathbb{R}^{TN\times 1}$ from the union of $L$ subspaces of $Tk$ dimensional subspaces $\mathcal{A}$, with probability
%at least $1-e^{-t}$.

\textbf{Theorem 5.} \emph{Let a matrix $A \in \mathbb{R}^{TM\times TN} $ be composed by sub-Gaussian random matrices $\{A'_t \in \mathbb{R}^{M\times N} \}_{t=1}^T$ as in (\ref{eqn:bigA}). For any fixed subset $S\subset \{1,2,...,TN\}$ with $|S|=TK$ and $0<\delta<1$, we have with probability exceeding} $1-2(12/\delta)^{TK} e^{-c_1 \frac{M}{2} \min(c_2^2\delta^2 \Gamma_2, c_2\delta \Gamma_\infty)}$:
\begin{eqnarray}
(1-\delta)||X||_2 \leq ||A_S X||_2 \leq (1+\delta)||X||_2, \label{eqn:RIP2}
\end{eqnarray}
\emph{for all $X=[x_1;x_2;...;x_T] \in   \mathbb{R}^{Tk \times 1}$.
$c_1$ and $c_2$ are absolute constants, $\Gamma_2 = \frac{(\sum_{t=1}^T ||x_t||^2_2 )^2 }{\sum_{t=1}^T ||x_t||^4_2}$ and $\Gamma_\infty = \frac{\sum_{t=1}^T ||x_t||^2_2  }{\max_{t=1}^T ||x_t||^2_2}$.}  \vspace{0.25cm}

\emph{Proof}. Let's denote $\bar{X}= X/||X||_2$ and we have $||\bar{X}||_2=1$. We choose a finite set of points $Q=\{q_i\}$, such that $q_i\in   \mathbb{R}^{Tk \times 1}$ and  $||q_i||_2 = 1$ for all $i$. We have $\min_i ||\bar{X}-q_i||^2 \leq \epsilon_1$ and covering number satisfies $|Q| \leq (1+2/\epsilon_1)^{TK}$ for any $\epsilon_1 >0$ (see Chap 13 of \cite{lorentz1996constructive}).

As the block-diagonal matrix $A$ is composed by sub-Gaussian random matrices, we have for each $i$ and any $\epsilon_2 >0$:
\begin{align}
\mathrm{P}(| ||A q_i||_2^2 - ||q_i||_2^2 | \geq & \epsilon_2 ||q_i||_2^2 )  \notag \\ &\leq 2e^{-c_1 \frac{M}{2} \min(c_2^2\epsilon_2^2 \Gamma_2, c_2\epsilon_2 \Gamma_\infty)},
\end{align}
with $\Gamma_2$ and $\Gamma_\infty$ defined above. This probability is indicated in Theorem III.1 of \cite{park2011concentration}.

Taking union bound, we obtain with probability exceeding $1- 2(1+2/\epsilon_1)^{TK}e^{-c_1 \frac{M}{2} \min(c_2^2\epsilon_2^2 \Gamma_2, c_2 \epsilon_2 \Gamma_\infty)}$:
\begin{eqnarray}
(1-\epsilon_2)\leq ||A_S q_i||_2^2 \leq (1+\epsilon_2), \ \mathrm{for} \ \mathrm{all} \  q_i \in Q,
\end{eqnarray}
which gives
\begin{eqnarray}
(1-\epsilon_2) \leq ||A_S q_i||_2 \leq (1+\epsilon_2), \ \mathrm{for} \ \mathrm{all} \  q_i \in Q.
\end{eqnarray}

Now we define $\rho$ as the smallest nonnegative number such that
\begin{eqnarray}
{||A_S \bar{X}||_2} \leq (1+\rho), \label{eqn:1rho}
\end{eqnarray}
for all $\bar{X} \in   \mathbb{R}^{Tk \times 1}$ and $||\bar{X}||_2 =1$. We have
\begin{align}
{||A_S \bar{X}||_2} &\leq {||A_S q_i||_2 + ||A_s(\bar{X}-q_i)||_2 } \notag \\
&\leq {||A_S q_i||_2} + {||A_s(\bar{X}-q_i)||_2}   \notag \\
&\leq (1+\epsilon_2) + (1+\rho) \epsilon_1.
\end{align}

As $\rho$ as the smallest nonnegative number for (\ref{eqn:1rho}), we have:
\begin{align}
1 + \rho \leq (1+\epsilon_2) + (1+\rho) \epsilon_1,
\end{align}
and
\begin{align}
\rho \leq (\epsilon_1+\epsilon_2)/(1- \epsilon_1).
\end{align}

Note the above result holds for any $\epsilon_1$ and $\epsilon_2$. We choose $\epsilon_1= \delta/4 $ and $\epsilon_2= \delta/2$. Since $0<\delta<1$, it is easy to see that $\rho\leq \delta$, which proves
\begin{eqnarray}
{||A_S \bar{X}||_2} \leq (1+\delta). \label{eqn:rho}
\end{eqnarray}

Similar, ${(1-\delta) \leq||A_S \bar{X}||_2}$  can be proved using the same way. Finally, we obtain with probability exceeding $1- 2(12/\delta)^{TK}e^{-c_1 \frac{M}{2} \min(c_2^2 \delta^2 \Gamma_2, c_2 \delta \Gamma_\infty)}$:
\begin{eqnarray}
(1-\delta) \leq \frac{||A_S X||_2}{||X||_2} \leq (1+\delta), \label{eqn:rho}
\end{eqnarray}
which completes the proof as $1+2/\epsilon_1 = (\delta+8)/\delta \leq 12/\delta$.

Based on this theorem, we know that any $Tk$-sparse $X \in \mathbb{R}^{TN\times 1}$ satisfies
\begin{eqnarray}
(1-\delta)||X||_2 \leq ||A X||_2 \leq (1+\delta)||X||_2, \label{eqn:frip}
\end{eqnarray}
with probability exceeding $1-2(12/\delta)^{TK} e^{-c_1 \frac{M}{2} \min(c_2^2\delta^2 \Gamma_2, c_2\delta \Gamma_\infty)}$.

Suppose there are $L$ combinations of such set $S$, from appendix A and B we know that
\begin{eqnarray}
L \leq \left\{\begin{matrix}
\frac{e^k N}{k+1} & \textrm{if} \ k \leq \left \lfloor \log_2 N \right \rfloor ,\\
\frac{4^k (c_4 N)}{k} & \textrm{if} \ k>\left \lfloor \log_2 N \right
\rfloor .
\end{matrix}\right. \label{eqn:Ltree}
\end{eqnarray}
for forest sparse data.

By taking the union bound, we known that (\ref{eqn:RIP2}) fails with probability less than
\begin{eqnarray}
2L(12/\delta_{\mathcal{F}_{T,k}})^{TK} e^{-c_1 \frac{M}{2} \min(c_2^2\delta_{\mathcal{F}_{T,k}}^2 \Gamma_2, c_2\delta_{\mathcal{F}_{T,k}} \Gamma_\infty)} \leq e^{-t},
\label{eqn:pb}
\end{eqnarray}
which gives
%\begin{eqnarray}
%JM \geq \frac{2J}{c_1 \min(c_2^2\delta^2 \Gamma_2, c_2\delta \Gamma_\infty)} (\ln(2L) + TK\ln (\frac{12}{\delta})+ t)
%\label{eqn:fdbbound}
%\end{eqnarray}
\begin{eqnarray}
TM \geq \left\{\begin{matrix}
\ \frac{2T(  \ln 2 +  k + \ln(N/(k+1)) + Tk\ln(12/\delta_{\mathcal{F}_{T,k}})  +t)}{c_1 \min(c_2^2\delta_{\mathcal{F}_{T,k}}^2 \Gamma_2, c_2\delta_{\mathcal{F}_{T,k}} \Gamma_\infty)}, \vspace{0.15cm} \\  \qquad \textrm{if} \ k \leq \left \lfloor \log_2 N
\right \rfloor , \vspace{0.15cm} \\
\ \frac{2T(\ln2 + k\ln4 + \ln((c_3 N)/k) + Tk\ln(12/\delta_{\mathcal{F}_{T,k}})   +t)}{c_1 \min(c_2^2\delta_{\mathcal{F}_{T,k}}^2 \Gamma_2, c_2\delta_{\mathcal{F}_{T,k}} \Gamma_\infty)} ,
\vspace{0.15cm} \\ \qquad \textrm{if} \ k>\left \lfloor \log_2 N \right \rfloor .
\end{matrix}\right.
\end{eqnarray}
From this one theorem 4 can be easily derived.
For both cases, the bound can be written as $ TM =\mathcal{O}(\frac{T^2k+T\log(N/k)}{\min(\Gamma_2, \Gamma_\infty)})$.

% Can use something like this to put references on a page
% by themselves when using endfloat and the captionsoff option.
\ifCLASSOPTIONcaptionsoff
  \newpage
\fi

% You can push biographies down or up by placing
% a \vfill before or after them. The appropriate
% use of \vfill depends on what kind of text is
% on the last page and whether or not the columns
% are being equalized.

%\vfill

% Can be used to pull up biographies so that the bottom of the last one
% is flush with the other column.
%\enlargethispage{-5in}

\bibliographystyle{IEEEtran}
\bibliography{IEEEbib}

% Generated by IEEEtran.bst, version: 1.13 (2008/09/30)
\begin{thebibliography}{10}
\providecommand{\url}[1]{#1}
\csname url@samestyle\endcsname
\providecommand{\newblock}{\relax}
\providecommand{\bibinfo}[2]{#2}
\providecommand{\BIBentrySTDinterwordspacing}{\spaceskip=0pt\relax}
\providecommand{\BIBentryALTinterwordstretchfactor}{4}
\providecommand{\BIBentryALTinterwordspacing}{\spaceskip=\fontdimen2\font plus
\BIBentryALTinterwordstretchfactor\fontdimen3\font minus
  \fontdimen4\font\relax}
\providecommand{\BIBforeignlanguage}[2]{{%
\expandafter\ifx\csname l@#1\endcsname\relax
\typeout{** WARNING: IEEEtran.bst: No hyphenation pattern has been}%
\typeout{** loaded for the language `#1'. Using the pattern for}%
\typeout{** the default language instead.}%
\else
\language=\csname l@#1\endcsname
\fi
#2}}
\providecommand{\BIBdecl}{\relax}
\BIBdecl

\bibitem{candes2006robust}
E.~Cand{\`e}s, J.~Romberg, and T.~Tao, ``Robust uncertainty principles: Exact
  signal reconstruction from highly incomplete frequency information,''
  \emph{IEEE Trans. Inf. Theory}, vol.~52, no.~2, pp. 489--509, 2006.

\bibitem{donoho2006compressed}
D.~Donoho, ``Compressed sensing,'' \emph{IEEE Trans. Inf. Theory}, vol.~52,
  no.~4, pp. 1289--1306, 2006.

\bibitem{candes2006compressive}
E.~Cand{\`e}s, ``Compressive sampling,'' in \emph{Proc. Int. Congr. Math.},
  2006, pp. 1433--1452.

\bibitem{candes2007sparsity}
E.~Candes and J.~Romberg, ``Sparsity and incoherence in compressive sampling,''
  \emph{Inverse problems}, vol.~23, no.~3, p. 969, 2007.

\bibitem{candes2006stable}
E.~Candes, J.~Romberg, and T.~Tao, ``Stable signal recovery from incomplete and
  inaccurate measurements,'' \emph{Comm. Pure Appl. Math.}, vol.~59, no.~8, pp.
  1207--1223, 2006.

\bibitem{natarajan1995sparse}
B.~Natarajan, ``Sparse approximate solutions to linear systems,'' \emph{SIAM J.
  Sci. Comput.}, vol.~24, no.~2, pp. 227--234, 1995.

\bibitem{tibshirani1996regression}
R.~Tibshirani, ``Regression shrinkage and selection via the lasso,'' \emph{J.
  R. Stat. Soc. Series B Stat. Methodol.}, pp. 267--288, 1996.

\bibitem{chen1998atomic}
S.~Chen, D.~Donoho, and M.~Saunders, ``Atomic decomposition by basis pursuit,''
  \emph{SIAM J. Sci. Comput.}, vol.~20, no.~1, pp. 33--61, 1998.

\bibitem{donoho2003optimally}
D.~Donoho and M.~Elad, ``Optimally sparse representation in general
  (nonorthogonal) dictionaries via $\ell_1$ minimization,'' \emph{Optimally},
  vol. 100, no.~5, pp. 2197--2202, 2002.

\bibitem{tropp2004greed}
J.~Tropp, ``Greed is good: Algorithmic results for sparse approximation,''
  \emph{IEEE Trans. Inf. Theory}, vol.~50, no.~10, pp. 2231--2242, 2004.

\bibitem{needell2009cosamp}
D.~Needell and J.~Tropp, ``Co{S}a{MP}: Iterative signal recovery from
  incomplete and inaccurate samples,'' \emph{Appl. Computat. Harmon. Anal.},
  vol.~26, no.~3, pp. 301--321, 2009.

\bibitem{beck2009afast}
A.~Beck and M.~Teboulle, ``A fast iterative shrinkage-thresholding algorithm
  for linear inverse problems,'' \emph{SIAM J. Imaging Sci.}, vol.~2, no.~1,
  pp. 183--202, 2009.

\bibitem{figueiredo2007gradient}
M.~Figueiredo, R.~Nowak, and S.~Wright, ``Gradient projection for sparse
  reconstruction: Application to compressed sensing and other inverse
  problems,'' \emph{IEEE J. Sel. Topics Signal Process.}, vol.~1, no.~4, pp.
  586--597, 2007.

\bibitem{koh2007interior}
K.~Koh, S.~Kim, and S.~Boyd, ``An interior-point method for large-scale
  l1-regularized logistic regression,'' \emph{J. Mach. Learn. Res.}, vol.~8,
  no.~8, pp. 1519--1555, 2007.

\bibitem{ji2008bayesian}
S.~Ji, Y.~Xue, and L.~Carin, ``Bayesian compressive sensing,'' \emph{IEEE
  Trans. Signal Process.}, vol.~56, no.~6, pp. 2346--2356, 2008.

\bibitem{donoho2009message}
D.~Donoho, A.~Maleki, and A.~Montanari, ``Message-passing algorithms for
  compressed sensing,'' in \emph{Proc. National Academy of Sciences}, vol. 106,
  no.~45, 2009, pp. 18\,914--18\,919.

\bibitem{huang2011learning}
J.~Huang, T.~Zhang, and D.~Metaxas, ``Learning with structured sparsity,''
  \emph{J. Mach. Learn. Res.}, vol.~12, pp. 3371--3412, 2011.

\bibitem{baraniuk2010model}
R.~Baraniuk, V.~Cevher, M.~Duarte, and C.~Hegde, ``Model-based compressive
  sensing,'' \emph{IEEE Trans. Inf. Theory}, vol.~56, no.~4, pp. 1982--2001,
  2010.

\bibitem{huang2011structured}
J.~Huang, ``Structured sparsity: theorems, algorithms and applications,'' Ph.D.
  dissertation, Rutgers University, 2011.

\bibitem{meng2011collaborative}
J.~Meng, W.~Yin, H.~Li, E.~Hossain, and Z.~Han, ``Collaborative spectrum
  sensing from sparse observations in cognitive radio networks,'' \emph{IEEE J.
  Sel. Areas Commun.}, vol.~29, no.~2, pp. 327--337, 2011.

\bibitem{krim1996two}
H.~Krim and M.~Viberg, ``Two decades of array signal processing research: the
  parametric approach,'' \emph{IEEE Signal Process. Mag.}, vol.~13, no.~4, pp.
  67--94, 1996.

\bibitem{baron2005distributed}
D.~Baron, M.~Wakin, M.~Duarte, S.~Sarvotham, and R.~Baraniuk, ``Distributed
  compressed sensing,'' \emph{arXiv preprint arXiv:0901.3403}, 2005.

\bibitem{majumdar2010compressive}
A.~Majumdar and R.~Ward, ``Compressive color imaging with group-sparsity on
  analysis prior,'' in \emph{Proc. IEEE Int. Conf. Image Process.(ICIP)}, 2010,
  pp. 1337--1340.

\bibitem{bilgic2011multi}
B.~Bilgic, V.~Goyal, and E.~Adalsteinsson, ``Multi-contrast reconstruction with
  bayesian compressed sensing,'' \emph{Magn. Reson. Med.}, vol.~66, no.~6, pp.
  1601--1615, 2011.

\bibitem{huang2012fast}
J.~Huang, C.~Chen, and L.~Axel, ``Fast {M}ulti-contrast {MRI}
  {R}econstruction,'' in \emph{Proc. Med. Image Comput. Comput. Assist. Interv.
  (MICCAI)}, 2012, pp. 281--288.

\bibitem{huang2010benefit}
J.~Huang and T.~Zhang, ``The benefit of group sparsity,'' \emph{Ann. Stat.},
  vol.~38, no.~4, pp. 1978--2004, 2010.

\bibitem{huang2009learningd}
J.~Huang, X.~Huang, and D.~Metaxas, ``Learning with dynamic group sparsity,''
  in \emph{Proc. Int. Conf. Comput. Vis. (ICCV)}, 2009, pp. 64--71.

\bibitem{yuan2005model}
M.~Yuan and Y.~Lin, ``Model selection and estimation in regression with grouped
  variables,'' \emph{J. R. Stat. Soc. Series B Stat. Methodol.}, vol.~68,
  no.~1, pp. 49--67, 2005.

\bibitem{bach2008consistency}
F.~Bach, ``Consistency of the group lasso and multiple kernel learning,''
  \emph{J. Mach. Learn. Res.}, vol.~9, pp. 1179--1225, 2008.

\bibitem{cotter2005sparse}
S.~Cotter, B.~Rao, K.~Engan, and K.~Kreutz-Delgado, ``Sparse solutions to
  linear inverse problems with multiple measurement vectors,'' \emph{IEEE
  Trans. Signal Process.}, vol.~53, no.~7, pp. 2477--2488, 2005.

\bibitem{van2008probing}
E.~Van Den~Berg and M.~Friedlander, ``Probing the pareto frontier for basis
  pursuit solutions,'' \emph{SIAM J. Sci. Comput.}, vol.~31, no.~2, pp.
  890--912, 2008.

\bibitem{deng2011group}
W.~Deng, W.~Yin, and Y.~Zhang, ``Group sparse optimization by alternating
  direction method,'' \emph{TR11-06, Department of Computational and Applied
  Mathematics, Rice University}, 2011.

\bibitem{wipf2007empirical}
D.~Wipf and B.~Rao, ``An empirical bayesian strategy for solving the
  simultaneous sparse approximation problem,'' \emph{IEEE Trans. Signal
  Process.}, vol.~55, no.~7, pp. 3704--3716, 2007.

\bibitem{ji2009multitask}
S.~Ji, D.~Dunson, and L.~Carin, ``Multitask compressive sensing,'' \emph{IEEE
  Trans. Signal Process.}, vol.~57, no.~1, pp. 92--106, 2009.

\bibitem{ziniel2011efficient}
J.~Ziniel and P.~Schniter, ``Efficient high-dimensional inference in the
  multiple measurement vector problem,'' \emph{arXiv preprint arXiv:1111.5272},
  2011.

\bibitem{manduca1996wavelet}
A.~Manduca and A.~Said, ``Wavelet compression of medical images with set
  partitioning in hierarchical trees,'' in \emph{Proc. Int. Conf. IEEE
  Engineering in Medicine and Biology Soc. (EMBS)}, vol.~3, 1996, pp.
  1224--1225.

\bibitem{he2009exploiting}
L.~He and L.~Carin, ``Exploiting structure in wavelet-based bayesian
  compressive sensing,'' \emph{IEEE Trans. Signal Process.}, vol.~57, no.~9,
  pp. 3488--3497, 2009.

\bibitem{som2012compressive}
S.~Som and P.~Schniter, ``Compressive imaging using approximate message passing
  and a markov-tree prior,'' \emph{IEEE Trans. Signal Process.}, vol.~60,
  no.~7, pp. 3439--3448, 2012.

\bibitem{rao2011convex}
N.~Rao, R.~Nowak, S.~Wright, and N.~Kingsbury, ``Convex approaches to model
  wavelet sparsity patterns,'' in \emph{Proc. IEEE Int. Conf. Image
  Process.(ICIP)}, 2011, pp. 1917--1920.

\bibitem{chen2012compressive}
C.~Chen and J.~Huang, ``Compressive {S}ensing {MRI} with {W}avelet {T}ree
  {S}parsity,'' in \emph{Proc. Adv. Neural Inf. Process. Syst. (NIPS)}, 2012,
  pp. 1124--1132.

\bibitem{kim2012tree}
S.~Kim and E.~Xing, ``Tree-guided group lasso for multi-response regression
  with structured sparsity, with an application to e{QTL} mapping,'' \emph{Ann.
  Appl. Stat.}, vol.~6, no.~3, pp. 1095--1117, 2012.

\bibitem{la2006tree}
C.~La and M.~Do, ``Tree-based orthogonal matching pursuit algorithm for signal
  reconstruction,'' in \emph{Proc. IEEE Int. Conf. Image Process.(ICIP)}, 2006,
  pp. 1277--1280.

\bibitem{chen2013benefit}
C.~Chen and J.~Huang, ``The benefit of tree sparsity in accelerated {MRI},''
  \emph{Med. Image Anal.}, 2013.

\bibitem{jacob2009group}
L.~Jacob, G.~Obozinski, and J.~Vert, ``Group lasso with overlap and graph
  lasso,'' in \emph{Proc. Int. Conf. Mach. Learn. (ICML)}, 2009, pp. 433--440.

\bibitem{sprechmann2011c}
P.~Sprechmann, I.~Ramirez, G.~Sapiro, and Y.~C. Eldar, ``C-hilasso: A
  collaborative hierarchical sparse modeling framework,'' \emph{IEEE Trans.
  Signal Process.}, vol.~59, no.~9, pp. 4183--4198, 2011.

\bibitem{mendelson2008uniform}
S.~Mendelson, A.~Pajor, and N.~Tomczak-Jaegermann, ``Uniform uncertainty
  principle for bernoulli and subgaussian ensembles,'' \emph{Constructive
  Approximation}, vol.~28, no.~3, pp. 277--289, 2008.

\bibitem{baraniuk2008simple}
R.~Baraniuk, M.~Davenport, R.~DeVore, and M.~Wakin, ``A simple proof of the
  restricted isometry property for random matrices,'' \emph{Constructive
  Approximation}, vol.~28, no.~3, pp. 253--263, 2008.

\bibitem{blumensath2009sampling}
T.~Blumensath and M.~Davies, ``Sampling theorems for signals from the union of
  finite-dimensional linear subspaces,'' \emph{IEEE Trans. Inf. Theory},
  vol.~55, no.~4, pp. 1872--1882, 2009.

\bibitem{chen2010graph}
X.~Chen, S.~Kim, Q.~Lin, J.~G. Carbonell, and E.~P. Xing, ``Graph-structured
  multi-task regression and an efficient optimization method for general fused
  lasso,'' \emph{arXiv preprint arXiv:1005.3579}, 2010.

\bibitem{chen2012two}
X.~Chen, X.~Shi, X.~Xu, Z.~Wang, R.~Mills, C.~Lee, and J.~Xu, ``A two-graph
  guided multi-task lasso approach for eqtl mapping,'' in \emph{Proc. Int.
  Conf. Artif. Intell. Stat.(AISTATS)}, 2012, pp. 208--217.

\bibitem{huang2011efficient}
J.~Huang, S.~Zhang, and D.~Metaxas, ``Efficient {MR} image reconstruction for
  compressed {MR} imaging,'' \emph{Med. Image Anal.}, vol.~15, no.~5, pp.
  670--679, 2011.

\bibitem{kowalski2012social}
M.~Kowalski, K.~Siedenburg, and M.~D{\"o}rfler, ``Social sparsity! neighborhood
  systems enrich structured shrinkage operators,'' \emph{IEEE Trans. Signal
  Process.}, vol.~61, no.~10, pp. 2498--2511, 2013.

\bibitem{huang2011composite}
J.~Huang, S.~Zhang, H.~Li, and D.~Metaxas, ``Composite splitting algorithms for
  convex optimization,'' \emph{Comput. Vis. Image Und.}, vol. 115, no.~12, pp.
  1610--1622, 2011.

\bibitem{lustig2007sparse}
M.~Lustig, D.~Donoho, and J.~Pauly, ``Sparse {MRI}: The application of
  compressed sensing for rapid {MR} imaging,'' \emph{Magn. Reson. Med.},
  vol.~58, no.~6, pp. 1182--1195, 2007.

\bibitem{rohlfing2009sri24}
T.~Rohlfing, N.~Zahr, E.~Sullivan, and A.~Pfefferbaum, ``The {SRI}24
  multichannel atlas of normal adult human brain structure,'' \emph{Hum. Brain
  Mapp.}, vol.~31, no.~5, pp. 798--819, 2009.

\bibitem{ma2008efficient}
S.~Ma, W.~Yin, Y.~Zhang, and A.~Chakraborty, ``An efficient algorithm for
  compressed {MR} imaging using total variation and wavelets,'' in \emph{Proc.
  IEEE Conf. Comput. Vis. Pattern Recognit. (CVPR)}, 2008, pp. 1--8.

\bibitem{yang2010fast}
J.~Yang, Y.~Zhang, and W.~Yin, ``A fast alternating direction method for
  {TVL1-L2} signal reconstruction from partial {F}ourier data,'' \emph{IEEE J.
  Sel. Topics Signal Process.}, vol.~4, no.~2, pp. 288--297, 2010.

\bibitem{majumdar2011joint}
A.~Majumdar and R.~Ward, ``Joint reconstruction of multiecho {MR} images using
  correlated sparsity,'' \emph{Magn. Reson. Imaging}, vol.~29, no.~7, pp.
  899--906, 2011.

\bibitem{chen2014exploiting}
C.~Chen and J.~Huang, ``Exploiting both intra-quadtree and inter-spatial
  structures for multi-contrast {MRI},'' in \emph{Proc. IEEE Int. Symp. Biomed.
  Imaging (ISBI)}, 2014.

\bibitem{pruessmann1999sense}
K.~Pruessmann, M.~Weiger, M.~Scheidegger, P.~Boesiger \emph{et~al.}, ``{SENSE}:
  sensitivity encoding for fast {MRI},'' \emph{Magn. Reson. Med.}, vol.~42,
  no.~5, pp. 952--962, 1999.

\bibitem{liang2009accelerating}
D.~Liang, B.~Liu, J.~Wang, and L.~Ying, ``Accelerating {SENSE} using compressed
  sensing,'' \emph{Magn. Reson. Med.}, vol.~62, no.~6, pp. 1574--1584, 2009.

\bibitem{majumdar2012calibration}
A.~Majumdar and R.~Ward, ``Calibration-less multi-coil {MR} image
  reconstruction,'' \emph{Magn. Reson. Imaging}, 2012.

\bibitem{chen2013calibrationless}
C.~Chen, Y.~Li, and J.~Huang, ``Calibrationless parallel {MRI} with joint total
  variation regularization,'' in \emph{Proc. Med. Image Comput. Comput. Assist.
  Interv. (MICCAI)}, 2013, pp. 106--114.

\bibitem{ma2009single}
J.~Ma, ``Single-pixel remote sensing,'' \emph{IEEE Geosci. Remote Sens. Lett.},
  vol.~6, no.~2, pp. 199--203, 2009.

\bibitem{lorentz1996constructive}
G.~G. Lorentz, M.~von Golitschek, and Y.~Makovoz, \emph{Constructive
  approximation: advanced problems}.\hskip 1em plus 0.5em minus 0.4em\relax
  Springer Berlin, 1996, vol. 304.

\bibitem{park2011concentration}
J.~Y. Park, H.~L. Yap, C.~J. Rozell, and M.~B. Wakin, ``Concentration of
  measure for block diagonal matrices with applications to compressive signal
  processing,'' \emph{IEEE Trans. Signal Process.}, vol.~59, no.~12, pp.
  5859--5875, 2011.

\end{thebibliography}

\end{document}